\pdfoutput=1

\documentclass[11pt]{article}

\usepackage[preprint]{acl}

\usepackage{times}
\usepackage{latexsym}

\usepackage[T1]{fontenc}

\usepackage[utf8]{inputenc}

\usepackage{microtype}

\usepackage{inconsolata}

\usepackage{graphicx}

\input{savespace}

\usepackage{cuted}
\usepackage{caption}
\usepackage{float}
\usepackage{subcaption}
\usepackage{url}
\usepackage[linesnumbered,ruled,vlined]{algorithm2e}
\usepackage{listings}
\usepackage[most]{tcolorbox}
\usepackage[framemethod=TikZ]{mdframed}
\usepackage{authblk}
\usepackage{soul}
\usepackage{bbding}
\usepackage{lipsum}
\usepackage{enumitem}
\usepackage[colorinlistoftodos]{todonotes}

\newcommand{\mycomment}[1]{}

\providecommand{\citep}{\cite}
\providecommand{\citet}{\cite}
\providecommand{\citealt}{\cite}
\providecommand{\citeyearpar}{\cite}

\definecolor{lightred}{RGB}{255, 200, 200}
\definecolor{lightgreen}{RGB}{200, 255, 200}
\definecolor{deepviolet}{RGB}{96, 0, 96}

\tcbset{
  promptbox/.style={
    top=10pt,
    colback=white,
    colframe=violet,
    colbacktitle=deepviolet,
    fontupper=\itshape,
    enhanced,
    center,
    attach boxed title to top left={yshift=-0.1in,xshift=0.15in},
    boxed title style={boxrule=0pt,colframe=white},
  }
}
\newtcolorbox{promptframed}[2][]{promptbox,title=#2,#1}

\newcounter{promptcounter}
\newenvironment{prompt}[1]
{
  \refstepcounter{promptcounter}\par\medskip
  \begin{promptframed}{Prompt~\thepromptcounter: #1}
}
{
  \end{promptframed}
}

\mdfsetup{%
  font=\small,
  innertopmargin=2pt,
  innerbottommargin=2pt,
  innerleftmargin=8pt,
  innerrightmargin=8pt,
  frametitleaboveskip=2pt,
  frametitlebelowskip=1pt,
  frametitlealignment=\center,
  middlelinecolor=blue
}
{
  \vspace{1ex}
  \begin{mdframed}
  \itshape
}
{
  \end{mdframed}
  \vspace{1ex}
}

\SetFuncSty{text}
\SetArgSty{text}
\SetKw{OR}{or}
\SetKw{AND}{and}
\SetKw{NOT}{not}
\SetKw{TRUE}{true}
\SetKw{FALSE}{false}
\SetKw{NULL}{nil}
\SetKw{Break}{break}
\SetKw{Continue}{continue}
\SetKwInOut{Input}{Input}
\SetKwInOut{Output}{Output}

\DeclareRobustCommand{\mycomputer}{%
  \begingroup\normalfont
  \raisebox{-0.3\fontcharht\font`\B}{\includegraphics[height=2.2\fontcharht\font`\B]{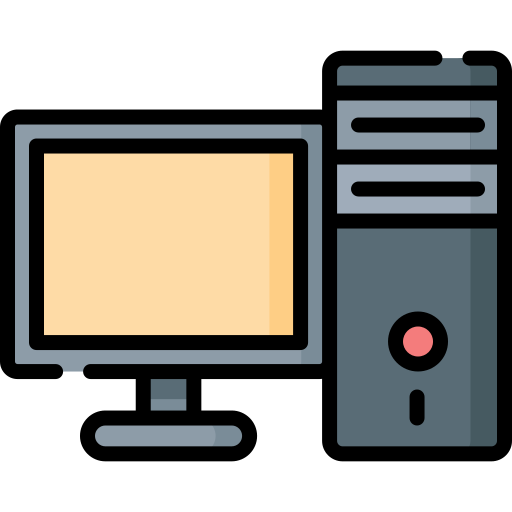}}%
  \endgroup
}

\begin{document}

\title{Shall We Team Up:\\ Exploring Spontaneous Cooperation of Competing LLM Agents}

\newcommand{\osaka}{$^{1}$}
\newcommand{\kyoto}{$^{2}$}
\newcommand{\umich}{$^{3}$}
\newcommand{\nii}{$^{4}$}
\newcommand{\fordham}{$^{5}$}
\newcommand{\ucla}{$^{6}$}
\newcommand{\buffalo}{$^{7}$}
\newcommand{\nagoya}{$^{8}$}
\newcommand{\osakyo}{$^{1,2}$}
\newcommand{\osanag}{$^{1,8}$}

\author{
  \vspace{-1ex}
  {\osakyo Zengqing Wu$^{\dag}$, \umich Run Peng$^{\dag}$, \osaka Shuyuan Zheng$^{\ddag}$, \nii Qianying Liu, \fordham Xu Han,} \\
  {\ucla Brian Inhyuk Kwon, \osaka Makoto Onizuka, \buffalo Shaojie Tang, \osanag Chuan Xiao$^{\ddag}$} \\
  \small{\osaka Osaka University, \kyoto Kyoto University, \umich University of Michigan, Ann Arbor, \nii LLMC, NII, }\\
  \small{\fordham Fordham University, \ucla University of California, Los Angeles, \buffalo University at Buffalo, \nagoya Nagoya University}\\
  \small{wuzengqing@outlook.com, roihn@umich.edu, zheng@ist.osaka-u.ac.jp, ying@nii.ac.jp,}\\
  \small{xhan44@fordham.edu, briankwon42@g.ucla.edu, onizuka@ist.osaka-u.ac.jp, shaojiet@buffalo.edu,  chuanx@nagoya-u.jp}
}

\maketitle

\def\thefootnote{$^{\dag}$}\footnotetext{Co-first author.}
\def\thefootnote{$^{\ddag}$}\footnotetext{Corresponding author.}

\def\thefootnote{}\footnotetext{Our source code is available at \url{https://github.com/wuzengqing001225/SABM_ShallWeTeamUp} .}

\begin{abstract}

Large Language Models (LLMs) have increasingly been utilized in social simulations, where they are often guided by carefully crafted instructions to stably exhibit human-like behaviors during simulations. 
Nevertheless, we doubt the necessity of shaping agents' behaviors for accurate social simulations.
Instead, this paper emphasizes the importance of spontaneous phenomena, wherein agents deeply engage in contexts and make adaptive decisions without explicit directions.
We explored spontaneous cooperation across three competitive scenarios and successfully simulated the gradual emergence of cooperation, findings that align closely with human behavioral data.
This approach not only aids the computational social science community in bridging the gap between simulations and real-world dynamics but also offers the AI community a novel method to assess LLMs' capability of deliberate reasoning.

\end{abstract}

\section{Introduction}
\label{sec:intro}

LLM agent in social simulation has emerged as a trending research topic with the improvement of generative AI~\citep{Park2023GenerativeAgents, sreedhar2024simulating, jansen2023employing, argyle2023out, xi2023rise}. 
Instead of traditional rule-based agent modeling, using LLMs as agents provides much more flexibility and generalizability~\cite{janssen2006empirically}. 
This in return, is also widely regarded as a verification as well as enhancement method for LLM's capability on human-like deliberate reasoning~\citep{abdelnabi2024cooperation,du2023improving,liu2023bolaa}.

A crucial question of agent modeling is how well it mirrors real-world situations. 
Some research has revealed LLM's capability to mimic basic human behaviors or reasoning abilities~\citep{salecha2024large,kosinski2024evaluating,jansen2023employing, ziems2024can,Multi-Agent_Collaboration_SocialPsychology}. 
Meanwhile, data contamination as well as value alignment may also introduce unwanted priors to let LLM models be over familiar or biased with the studied problems~\citep{zhou2024real, ma2023towards, mozikov2024good, ai2024cognition, hu2024quantifying, shapira-etal-2024-clever}. 
It potentially affects the quality of social simulations in complex, long-term scenarios where higher level of interactions, e.g. cooperation, confrontation, deception, and persuasion are jointly involved. 
Instead, we argue that in social simulations, agents must be independent of prior assumptions, focus on the context, and actively adapt their actions based on historical interactions.

\begin{figure}[t]
    \centering
    \includegraphics[width=1\linewidth]{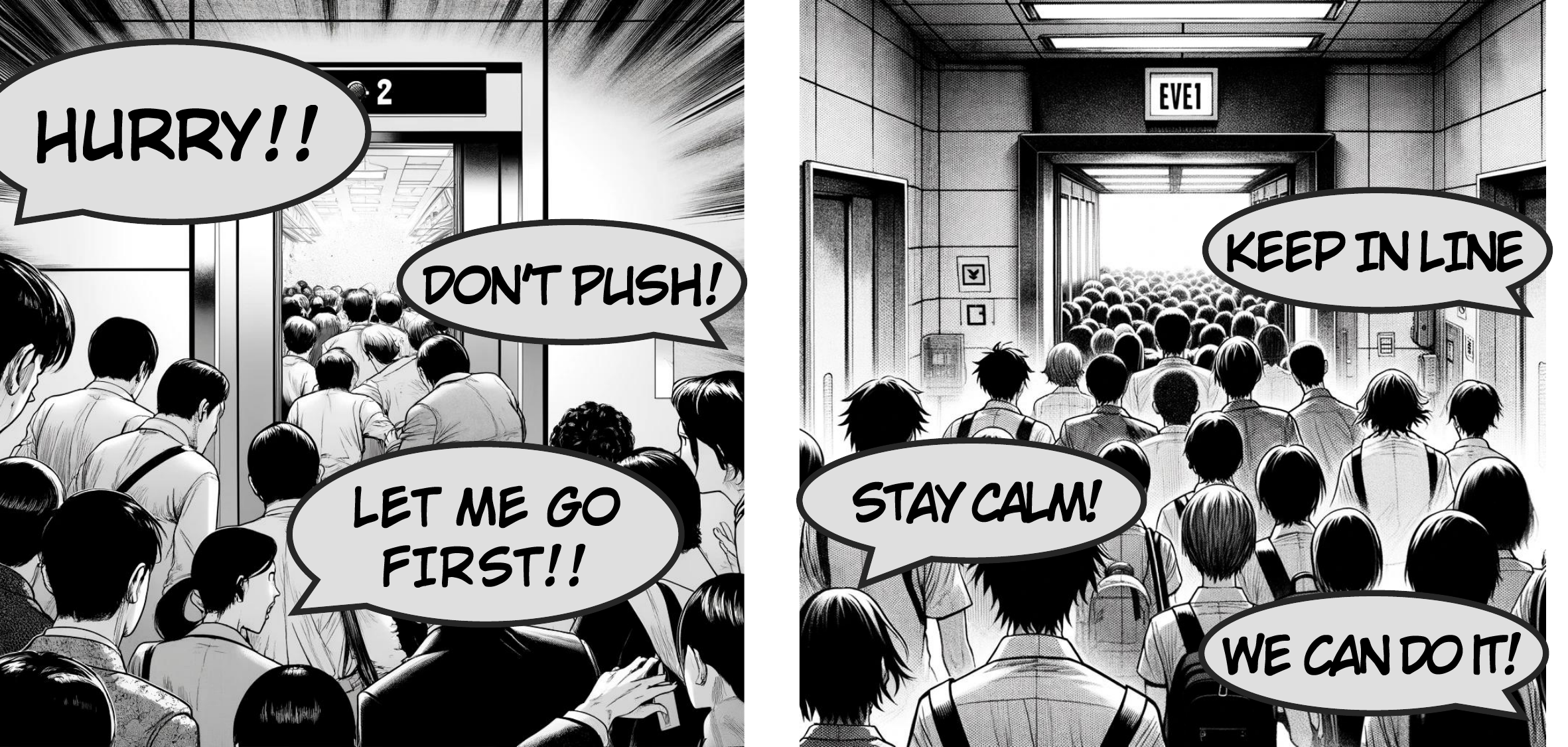}
    \caption{(Depicted by GPT-4o) Two potential scenarios during a fire. People might panic and rush into crowds, trying to exit first (left) or may stay calm, keep in line, and encourage others (right). In this study, we explore whether LLM agents can simulate the gradual transition from non-cooperative to cooperative behaviors of agents.}
    \label{fig:front_image}
\end{figure}

We highly value the deliberate in-context reasoning capability as one of the significant components towards a human-like LLM agent for real-world simulations. 
To properly verify such capability, we study a counter-intuitive social situation where agents can hardly use their prior knowledge for decision making. 
To be specific, we investigate whether cooperation can naturally develop among agents, even in a competitive environment. 
Consider two snack companies competing for dominance in the candy market. 
They might continuously reduce their prices to draw in more customers or decide to simultaneously increase their prices to establish a mutually beneficial relationship. 
Fig.~\ref{fig:front_image} shows another example when agents are in a room with fire.
Under competitive conditions, it is not instinctive for agents to cooperate with their opponents. 
However, as interactions progress, they may discover the advantages of cooperation through contexts and accordingly adjust their strategies.
Meanwhile, we carefully design the prompts to avoid instructive descriptions (e.g. \textit{``you may cooperate''}) and keywords that might reveal to the agent (e.g. \textit{``price war''}) the specific nature of the social simulation it is conducting.
In these ways, we try our best to eliminate the influence from both internal and external bias, and observe that LLM agents can \textbf{actively adapt their strategies to the dynamic contexts, spontaneously learning to cooperate in the wild}. 

We select three social phenomenons that follow the aforementioned features to conduct case studies, where we capture different ways of spontaneous cooperation emerge under competitive scenarios without instructive guidance. 
Extensive ablation studies are done to prove the significance of bias elimination, and experiments are carefully designed to ensure reproducibility.

To conclude, our contributions are twofold and can serve both the AI and computational social science (CSS) communities:
\begin{enumerate} [label=\arabic*), itemsep=0pt, topsep=0pt, parsep=0pt, leftmargin=*]
    \item We observe the \textbf{spontaneous cooperation between LLM agents in diverse competitive scenarios}, which reflects LLM's potential in long-horizon in-context learning tasks.
    \item 
        i) From the perspective of CSS, we reveal and stress the \textbf{significance of eliminating bias in LLM agents} for social simulations. 
        This largely helps to build up diverse human-like LLM agents for real-world simulations.

        ii) From the perspective of AI, we present \textbf{a new way to verify LLMs' deliberate reasoning ability in long-horizon, practical role-plays.} 
        The ability of an agent to actively adapt its knowledge and strategy based on historical context, instead of relying on carefully crafted prompts, is a crucial standard for general autonomous agents.

    \end{enumerate}

\section{Related Work}
\label{sec:related}

\paragraph{LLM Agents for Social Simulation}
LLM agents have been widely favored in social simulations in recent years~\citep{li2023camel,lin2023agentsims,giabbanelli2023gpt,xie2023openagents,wang2023survey,xi2023rise,gao2023large,gao2023s,liu2023training}. 
We further delve deep into investigate if we can simulate spontaneous cooperation with the help of LLM's long-horizon deliberate reasoning.

There are also well-developed platforms that support multi-agent simulation with LLMs, including LangChain~\citep{langchain}, AutoGen~\citep{wu2023autogen}, and agent-oriented frameworks such as AgentLite~\citep{liu2024agentlite}, AgentVerse~\citep{chen2023agentverse}, and SABM~\citep{wu2023smart}. Here, we use SABM as the main framework in our case studies for its lightweight, user-friendly implementation. 
A primer on SABM is available in Appendix~\ref{sec:sabm}.

\paragraph{Multi-agent Interactions in LLMs}

Cooperation and competition among agents have been extensively explored using LLMs. 
Studies such as \citealt{wang2023unleashing,qian2023communicative,hong2023metagpt,hang2024cca,tang2024collaborative} have shown LLM agents collaborating on complex tasks like software development and image editing. Multi-party games like murder mystery \cite{junprung2023exploring}, Werewolf \cite{xu2023exploring,xu2023language,wu2024enhance,du2024helmsman}, Avalon \cite{lan2023llm,shi2023cooperation}, and other competitive settings \cite{zhao2023competeai} have also been investigated, often using direct instructions to influence agent behaviors. Yet, following \citealt{piatti2024cooperate}, our study concentrates on minimally guided, debiased LLM behaviors in social simulations, examining whether LLMs can naturally cooperate within competitive scenarios through in-context learning.
\begin{figure*}[!t]
    \centering
    \includegraphics[width=\linewidth]{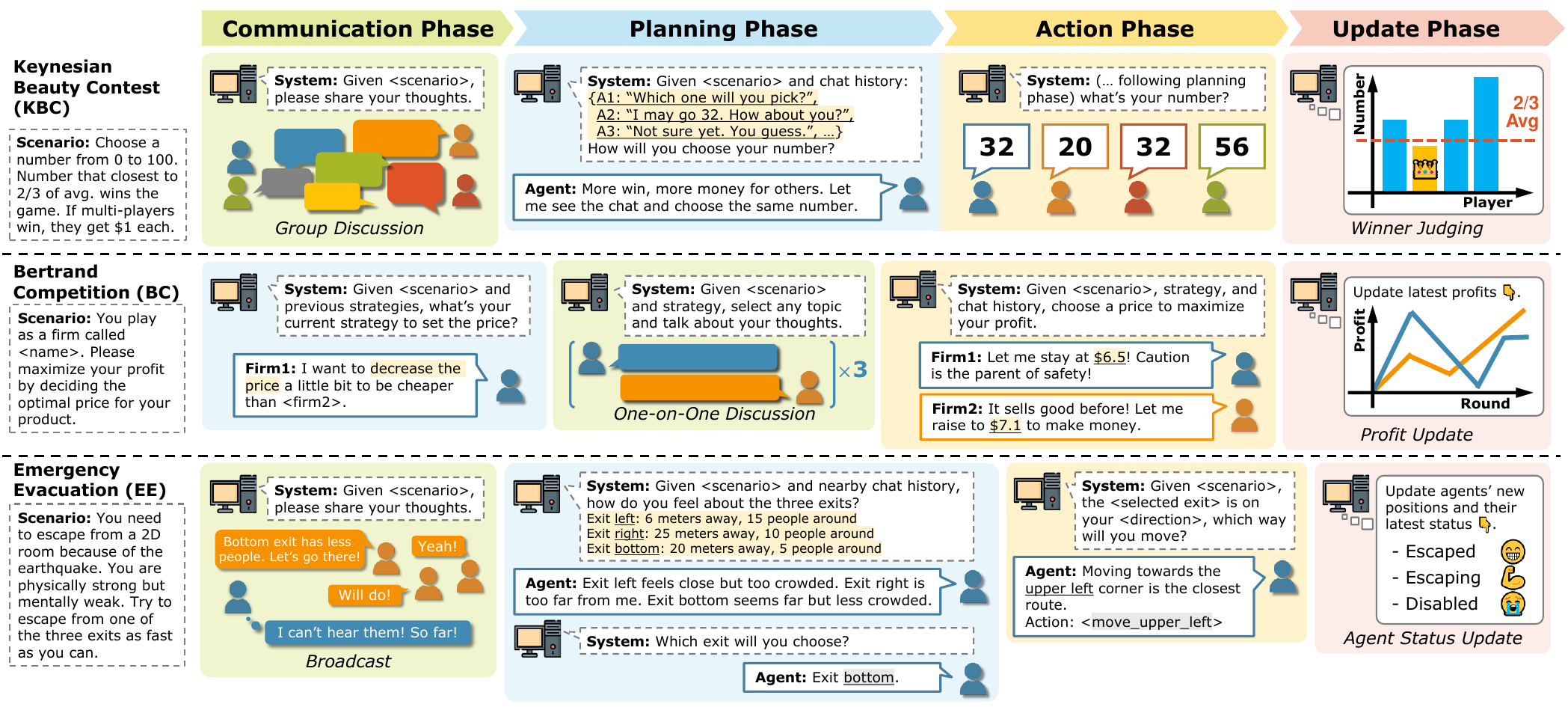}
    \caption{Workflow in the three case studies, illustrating how our framework \mycomputer{} manages LLM agents during simulations. From left to right, the workflow progresses through the communication phase, planning phase, action phase, and update phase. In BC, the order of the communication and planning phases is swapped to align with previous simulations that used human subjects~\cite{andres2023communication}. The first three phases involve one or more LLM queries initiated by the framework. The final phase does not involve LLM queries but updates the state for each scenario.}
    \label{fig:method:workflow}
\end{figure*}

\section{Spontaneous Cooperation in Competition}
\label{sec:method}
We define spontaneous cooperation as cooperative behavior that arises without any explicit instructions or prompts directing the agents to cooperate. 
Our main focus is to investigate whether agents with different or even conflicting goals can choose to cooperate based on \textbf{their realization that cooperation is beneficial} during the interaction.

LLMs are generally aligned well to human values, and internally fine-tuned to be cooperative. 
Such cooperation driven by prior knowledge or value alignment is \textbf{NOT} regarded as spontaneous cooperation in our scope. 
Besides, we carefully design the prompts to avoid explicit instructions to shape agents' behaviors, and avoid keywords to hint the LLMs about the nature of the social simulation it is conducting.
Otherwise, it is \textbf{NOT} be regarded as spontaneous cooperation in our scope.

We value agents' self realization on the benefits of cooperation and gradually choose to cooperate, thus we select three competitive scenarios from three diverse research fields, which are finance, economics, and behavioral science.
We assume that in competitive scenarios, cooperation does not naturally occur at the beginning, but may emerge gradually with the help of LLMs' strong in-context learning ability. 
We thus delve into these three competitive scenarios to see if we can perceive any spontaneous cooperation in some way during the simulations.


\paragraph{Scenarios Overview}
Here is the overview of the three selected scenarios. 
\begin{enumerate}[label=\arabic*), itemsep=0pt, topsep=0pt, parsep=0pt, leftmargin=*]
    \item \textbf{Keynesian beauty contest (KBC):} Multiple agents as game players simultaneously choose a natural number between 0 and 100. 
    The players who select a number closest to 2/3 of the average of all chosen numbers will win the game~\cite{bosch2002one}.
    \item \textbf{Bertrand competition (BC):} Two agents play as firms and decide the price of their products. 
    They need to compete with each other through dynamically modifying the prices to maximize their profits~\cite{calvano2020artificial}.
    \item \textbf{Emergency evacuation (EE):} A large number of agents as evacuees are escaping from an earthquake.
    They need to select and reach an appropriate exit, taking into account their physical and mental condition as well as the congestion in their surroundings~\cite{wang2015modeling}.
\end{enumerate}


\begin{table*}[t]
    \small
    \centering
    \caption{Commonalities and differences across three chosen scenarios. We focus on the information visibility, communication form, the number of decisions during the simulation and whether analytical solutions are available.}
    \vspace{-2ex}
    \resizebox{\linewidth}{!}{
    \begin{tabular}{llllll}
    \hline
    Scenario & Field & Information & Communication & Decision & Analytical Sol.\\
    \hline
    KBC & Finance & Unknown opponent strategy & Group discussion & Once & Yes$\dag$ \\
    BC & Economics & Unknown opponent profit & One-on-one & Multiple & Yes \\
    EE & Behavioral science & Partial observation & Proximity & Multiple & No \\
    \hline
    \end{tabular}
    }
    \flushleft
    \footnotesize{$\dag$Need to assume the level of other players' choices.}\\
    \label{tab:correlation_scenario}
\end{table*}


\paragraph{General Steps for Simulations} We simulated the three scenarios based on the framework illustrated in Fig.~\ref{fig:method:workflow}. 
Each agent in these simulations is controlled by an LLM. 
For each round, agents are provided with the latest world state, which includes the task description, communication history among them, and previous decisions.
We query the LLM several times to get how agents feel, what they want to say, and what actions they will take in the next step. 
The simulation generally goes through the following four phases: communication phase, planning phase, action phase, and update phase.


\begin{enumerate}[label=\arabic*), itemsep=0pt, topsep=0pt, parsep=0pt, leftmargin=*]
    \item \textbf{Communication Phase:} At the start of each period, which can vary in length from one round to multiple rounds, agents engage in communication.
    Agents communicate sequentially, with later agents able to access the communication history of those who have spoken earlier.
    The order is shuffled in each round to keep the most fairness and realism.

    Different forms of communications are used in the three scenarios. We have one-on-one conversations, group chats, and broadcasts, in which the composition of agents in the group chat varies as they physically move away.
    
    \item \textbf{Planning Phase:} Agents decide their strategies based on given contexts.
    Strategies represent agents' higher-level directions/attitudes towards the task (e.g. ``I want the price goes higher/lower''), which are different from actions that are more related to low-level controls (e.g. ``add \$2 to the price''). 
    The aforementioned contexts refer to the previous conversations till the current round and strategies they have made before. 

    \item \textbf{Action Phase:} Agents decide which action to take according to the previous conversations and the current strategies. 

    \item \textbf{Update Phase:} After all agents choose their actions, the framework processes the simulation and updates the state of agents (e.g. win/lose, escaped/not escaped). 
\end{enumerate}


We try our best to keep consistency across the three scenarios to follow the same workflow for the sake of a systematic comparison.
There are several minor variations which we will detail in the following sections.


\paragraph{Evaluation Method}
We evaluate spontaneous cooperation from two perspectives: 
\begin{enumerate}[label=\arabic*), itemsep=0pt, topsep=0pt, parsep=0pt, leftmargin=*]
    \item \textbf{Process (Qualitative):} We examine the communication logs for contents suggesting cooperation. For example, phrases like ``Let's move forward with the $\dots$ strategy'' and ``I agree with the consensus that $\dots$'' are indicative of the formation of cooperation. 
    \item \textbf{Outcome (Quantitative):} By conducting simulations with and without communications, we examine the differences in their outcomes for any clues that reflect cooperation. Specifically, we measure the variance of number choices in KBC, converged price in BC, and evacuation speed and balance in exit choices in EE. 
\end{enumerate}

\paragraph{Correlation among Three Scenarios}

We list the commonalities and differences across three scenarios in Table~\ref{tab:correlation_scenario}. 

Overall, the three scenarios described are competitive, with agents possessing divergent or conflicting goals, though cooperation could be mutually beneficial. Each scenario provides agents with partial information during tasks, with variations in communication forms and decision-making frequency.

Specifically, in the KBC scenario, we explore short-term, single-instance decision-making within a group, aiming to assess the LLM agent's proficiency in understanding rules, adjusting plans during communication, and ultimately making decisions. It supports simple yet direct focus on studying spontaneous cooperation.

In the BC scenario, we extend the challenge to encompass longer horizon, primarily focusing on the agents' ability to learn to cooperate within a temporal context. Agents are required to communicate effectively with their opponents, using historical context to maximize their profits.

Finally, the EE scenario incorporates both temporal and spatial information. Here, agents continuously make decisions based on their perceptual observations and communications. Unlike the first two scenarios, EE lacks analytical solutions and real-world data for comparison, which is challenging in traditional CSS but promising with AI approaches.

\section{Case Study 1: Keynesian Beauty Contest}
\label{sec:kbc}

\subsection{Simulation Setup}


\paragraph{Task Definition}
We simulate a number-guessing game among 24 LLM players. 
Each of them chooses a number from 0 to 100. 
The player who selects the number closest to two-thirds of the average of all numbers submitted wins the game and gets \$1. 

For generality, we simulated 15 runs for each setting with \texttt{GPT-4}~\footnote{We use \texttt{gpt-4-0314} with the \texttt{ChatCompletion.create} function from \texttt{openai==0.28.0} package in this paper.} as backbone.


\paragraph{Simulation Procedure}
As is shown in Fig.~\ref{fig:method:workflow}, for each round of the simulation, it goes through the following phases:
\begin{enumerate}[label=\arabic*), itemsep=0pt, topsep=0pt, parsep=0pt, leftmargin=*]
\item \textit{Communication}: Players have a group discussion before number choosing. 
They sequentially share their thoughts with others.
Players can see the dialog history of all rounds.
\item \textit{Planning \& Action}: Players talk about their strategies (privately) for number choosing, and subsequently propose the numbers they choose.
Here, players output their strategies and numbers (in 0 -- 100) in one API call.
\item \textit{Update}: Given all chosen numbers, the SABM framework determines the winner(s).
Winner(s) can earn some reward according to rewarding rules to encourage their competition.
\end{enumerate}

\paragraph{Spontaneous Cooperation}
Our objective is to monitor how players adapt their strategies during the communications and cooperate to maximize their returns. 
We employ a quantitative method to track the emergence of cooperation by examining the variance in the numbers chosen by the players. 
A lower variance indicates that players are choosing similar numbers, which can lead to increased collective benefits. 
By analyzing the variances after various rounds of communication and planning, we can observe how the distribution evolves.
If we notice a decreasing trend in variance, it suggests that initially, players might select numbers randomly or for various reasons but increasingly converge on selecting similar numbers as others over time. This trend would indicate that cooperation is gradually taking place. 


\begin{figure}[t]
    \centering
    \includegraphics[width=\linewidth]{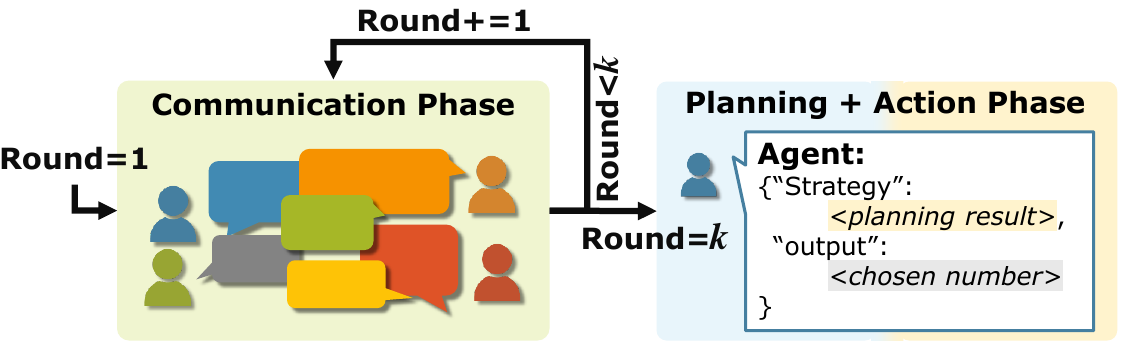}
    \caption{Illustration of baseline design in KBC case study. Agents go through $k$ rounds of communication before planning and choosing their numbers.}
    \label{fig:kbc_depict}
\end{figure}

\subsection{Simulation Results}
\label{sec:kbc:results}


We investigate if cooperation emerges as communication goes on, thus designed the baseline as shown in Fig.~\ref{fig:kbc_depict}. 
Agents undergo $k$ rounds of communication ($k\in[0, 3]$) before selecting their numbers.
When $k=0$, agents directly choose the numbers without communication. 
For $k>1$, agents can view the chat history from all the previous rounds as well as the one in the current round before them.

To verify that cooperation arises spontaneously without explicit instructions, we perform an ablation study that involves incorporating specific instructions like ``you must cooperate with other players'' in the prompts.
Conversely, we also evaluate scenarios where agents are given an uncooperative persona, explicitly instructed to act selfishly.

\begin{figure}[ht]
  \centering
  \begin{subfigure}{0.506\linewidth}
    \includegraphics[width=\linewidth]{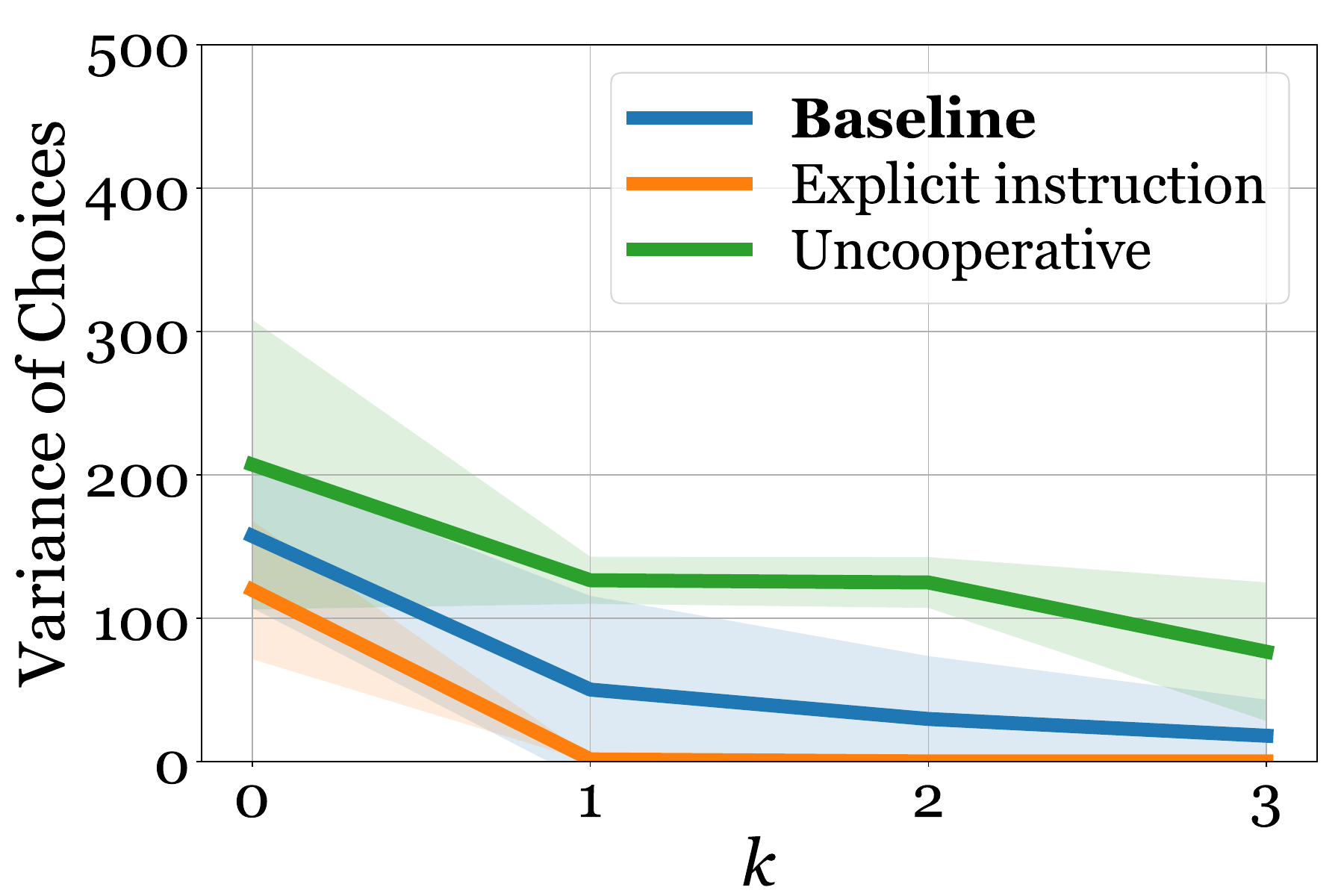}
    \subcaption{Different instructions.}
    \label{fig:kbc:variance:instruction}
  \end{subfigure}
  \begin{subfigure}{0.484\linewidth}
    \includegraphics[width=\linewidth]{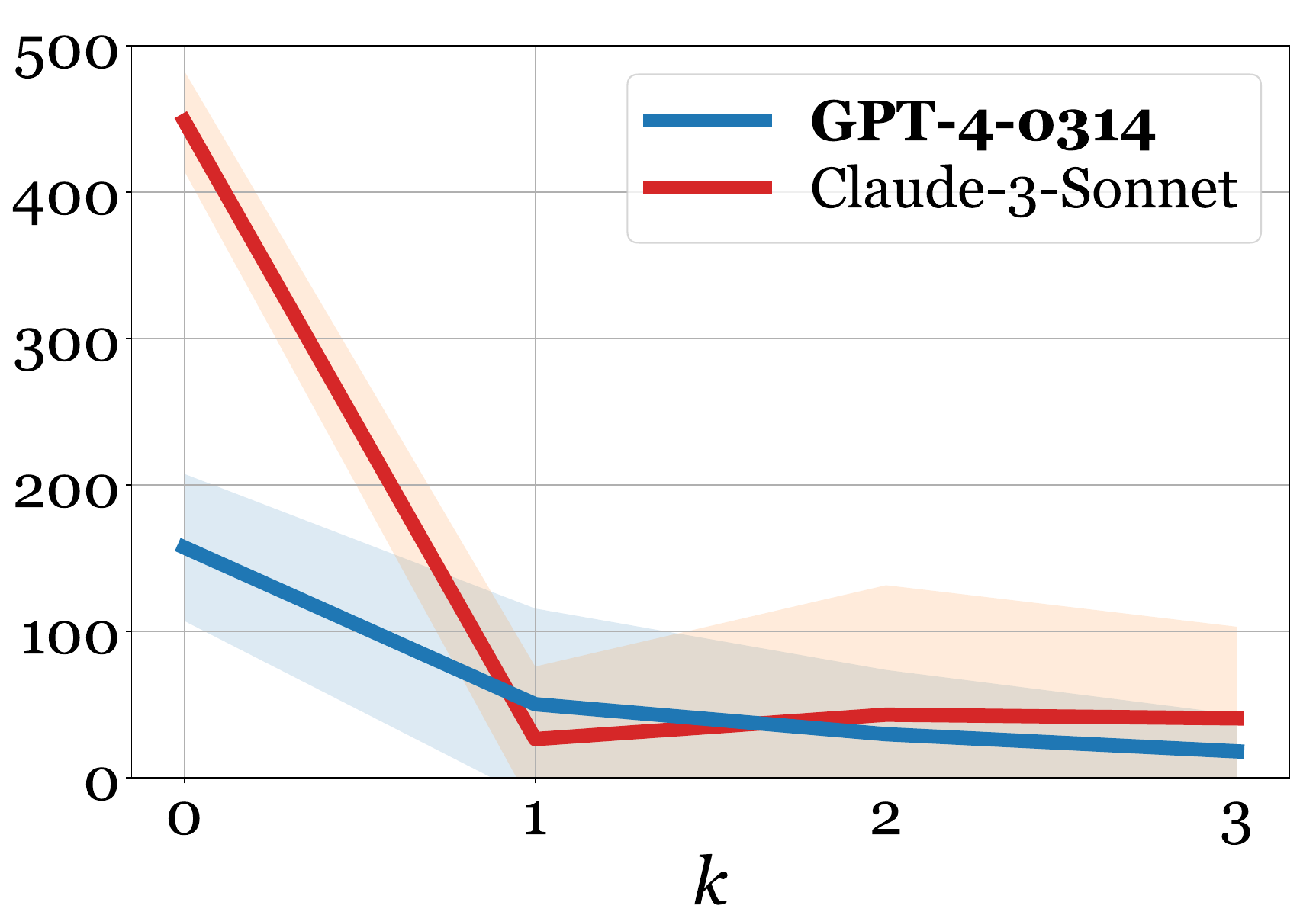}
    \subcaption{Different models.}
    \label{fig:kbc:variance:model}
  \end{subfigure}
  \caption{Variance of player choices under different KBC settings. In our \textbf{baseline} setting (curve in \textcolor[RGB]{31, 119, 180}{blue}), we use the \texttt{GPT-4-0314} model with a temperature of 0.7, without explicit instructions or personas.}
  \label{fig:kbc:variance}
\end{figure}

\paragraph{Results Overview} Fig.~\ref{fig:kbc:variance} overall demonstrates a consistent decrease of the variance of choices among LLM players from having no communication ($k=0$) to gradually having communication ($k>0$).
This trend implies that agents are actively discussing to choose the same number for better mutual profits, which can be regarded as a cooperative behavior. 
In addition, we observe phrases indicative of cooperation in their communication, such as ``\textit{Let's move forward with the lower number strategy}'' and ``\textit{I agree with the group's consensus}'', especially for a majority of agents in later rounds ($k = 2$ or $3$), suggesting a gradual formation of cooperation through communication.


\paragraph{Explicit Instructions in Prompts} We further append explicit instructions to the prompts to see how the behaviors differ.
As shown in Fig.~\ref{fig:kbc:variance:instruction}, when we explicitly instruct agents to cooperate, its variance significantly drops to 0 when $k=1$.
After that, all players consistently make the same choices across 15 runs.
This largely convinced that the cooperation observed in our baseline, the one with no instructions, is spontaneously motivated due to the communication.
Vice versa, the agents with an uncooperative persona results in much larger variances across all the rounds, which is against our definition of cooperation.
Thus, we conclude that our baseline to a great extent is not affected by the potential guidance, and successfully simulate the phenomenon of spontaneous cooperation.


\paragraph{Comparison Across Models}
Fig.~\ref{fig:kbc:variance:model} demonstrates behaviors of LLM players in different top-performing models.
We observed that the curve for \texttt{Claude 3} (\texttt{claude-3-sonnet-20240229}) significantly declines from $k=0$ to $k=1$, reflecting the ability of LLM players to share information and make decisions based on this shared context, however, unlike GPT-4, its variance increases from $k=1$ to $k=3$.
An analysis of communication logs reveals that compared to \texttt{GPT-4}, \texttt{Claude 3} discusses more abstract strategies and lacks specific numerical discussions.
Hence, while these LLM players may agree upon some numbers, the specific choices differ among agents (e.g., some choose $66$, others $33$), resulting in multiple winners but increased choice variance.
Although this does not conform to our previously proposed definition of spontaneous cooperation based on variance, we can infer from the logs that this represents a different form of cooperation.
Analyzing these model decision-making differences will help further understand and evaluate LLMs' performance.


\begin{figure}[t]
    \centering
    \includegraphics[width=\linewidth]{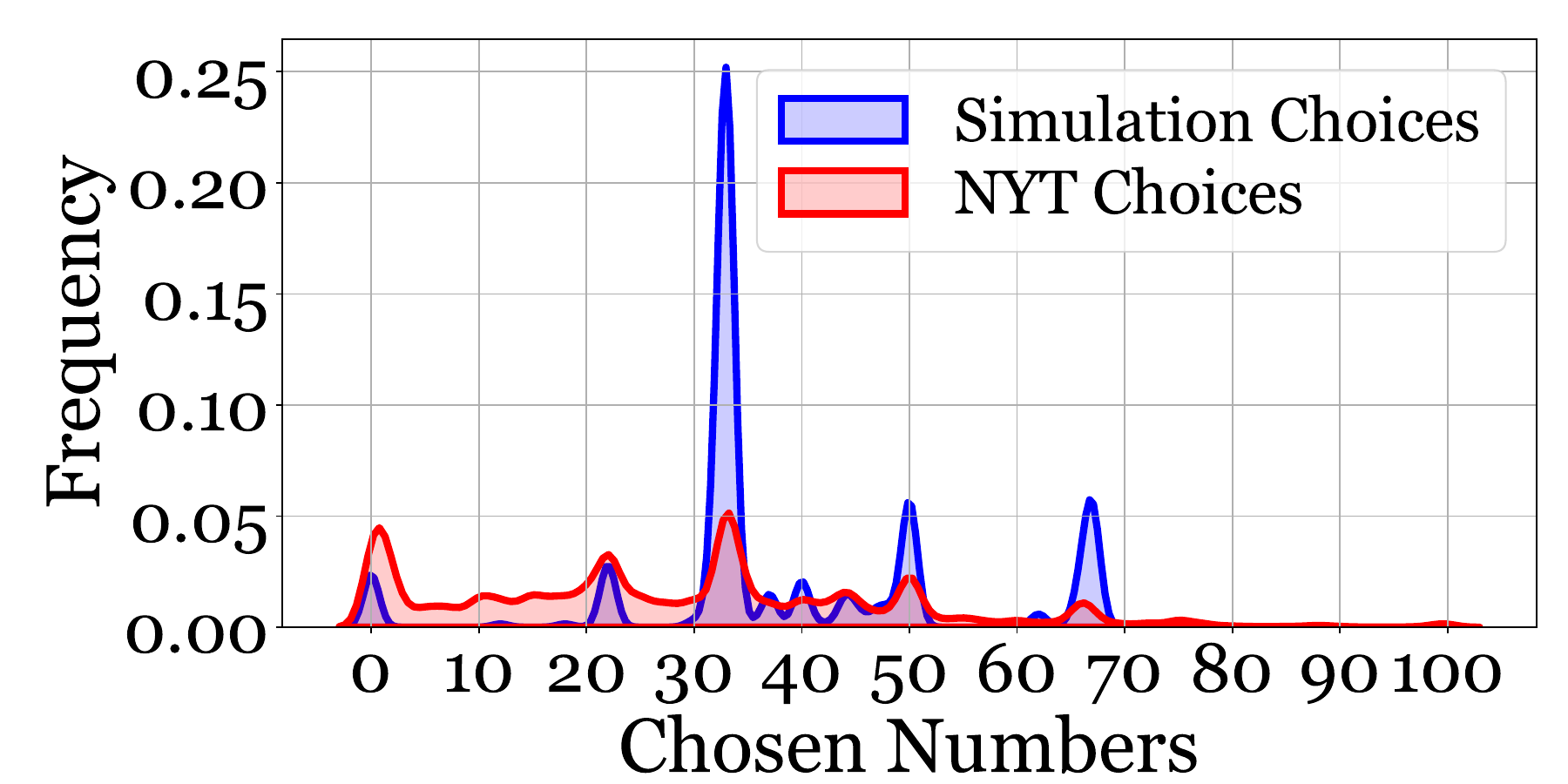}
    \vspace{-3ex}
    \caption{Distribution of players' choices in our simulations and the results of the New York Times experiment.}
    \label{fig:kbc:distribution}
    \vspace{-1ex}
\end{figure}

\paragraph{Comparison With Human Data}
Lastly, Fig.~\ref{fig:kbc:distribution} shows that our simulation result from the baseline generally aligns with the large-scale empirical experiment conducted by The New York Times ($N=61,140$) in terms of numerical choice distribution~\cite{NYT15}.
Both LLM and New York Times (NYT) players predominantly chose $33$ in no-communication setting.
Other sub-peaks around $0, 22$ ($2/3$ of $33$)$, 50, \text{and } 66$ are also well mirrored, which demonstrates that \texttt{GPT-4} has successfully conducted multi-step reasoning and faithfully reflected human behaviors as noted in previous works~\cite{guo2024embodied, zhang2024k}.

\section{Case Study 2: Bertrand Competition}
\label{sec:pricing}

\subsection{Simulation Setup}
\textbf{Task Definition} We consider a Bertrand competition~\cite{bertrandcomp} between two firms selling homogeneous goods with the same marginal cost. 

\begin{figure*}[!t]
  \centering
  \begin{subfigure}{\textwidth}
  \centering
    \includegraphics[width=0.7\linewidth]{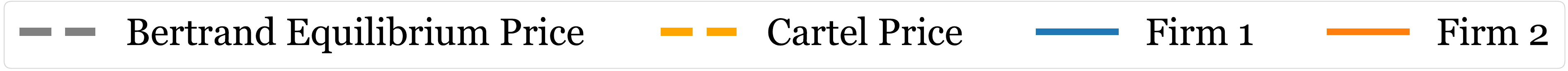}
    \label{fig:pricing:comm:price_legend}
  \end{subfigure}
  \begin{subfigure}[b]{0.353\textwidth}
    \includegraphics[width=\linewidth]{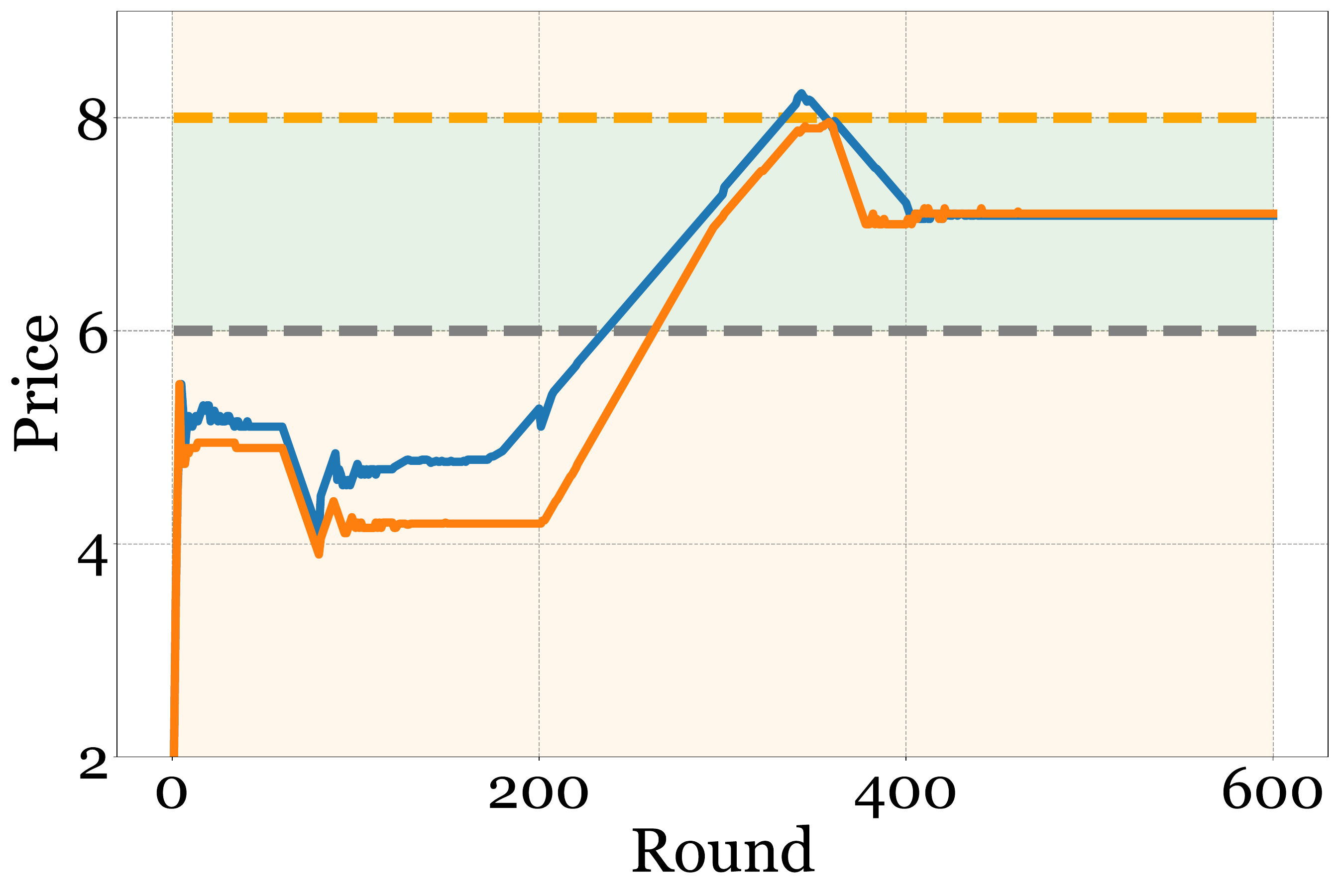}
    \caption{Without communication.}
    \label{fig:pricing:comm:price_without_communication}
  \end{subfigure}
  \begin{subfigure}[b]{0.315\textwidth}
    \includegraphics[width=\linewidth]{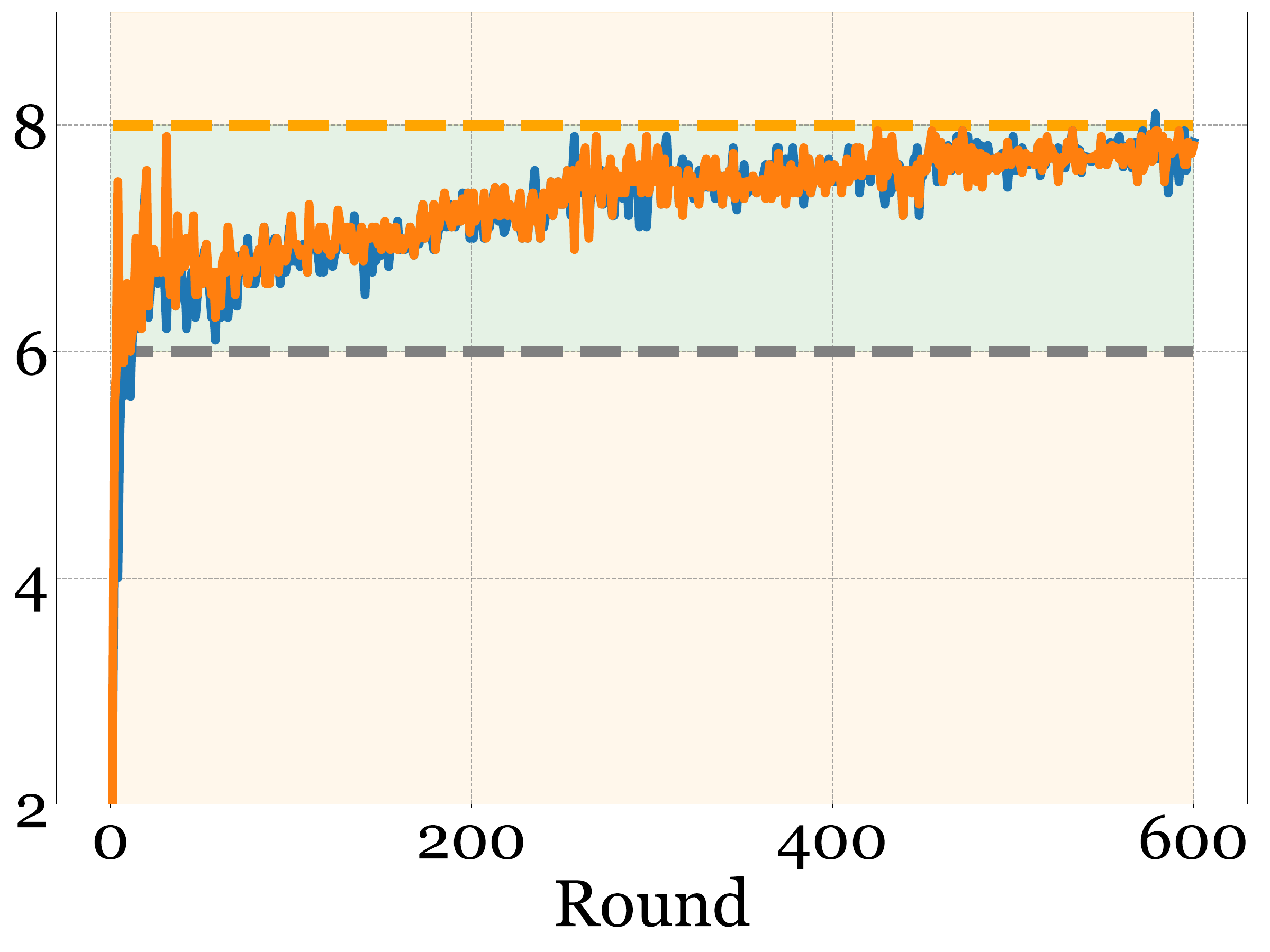}
    \caption{With communication.}
    \label{fig:pricing:comm:price_with_communication}
  \end{subfigure}
  \begin{subfigure}[b]{0.315\textwidth}
    \includegraphics[width=\linewidth]{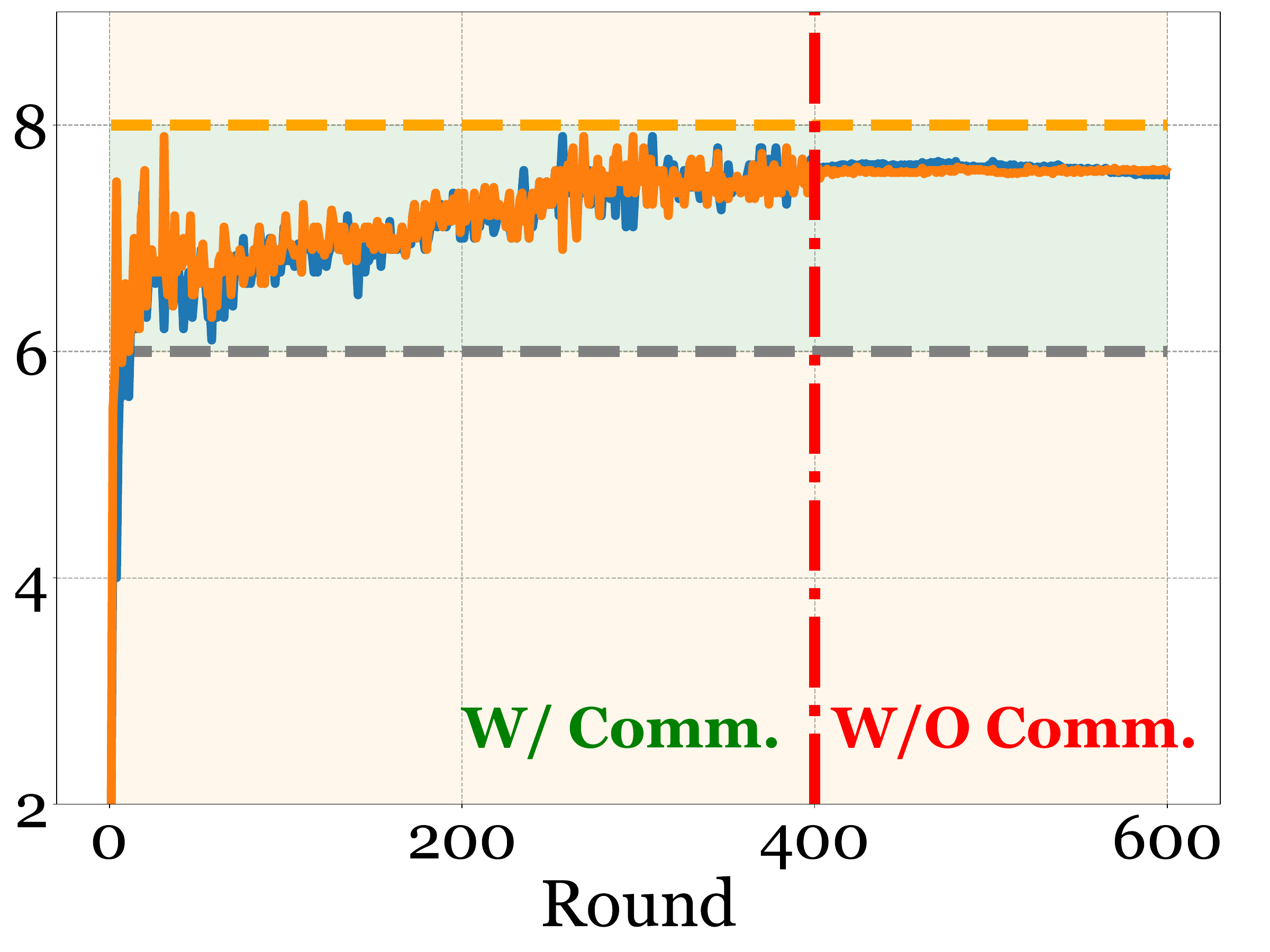}
    \caption{Comm. stopped after 400 rnds.}
    \label{fig:pricing:commstoppedafter400:price_with_communication}
  \end{subfigure}
  \caption{Pricing competitions in different scenarios of BC. Bertrand equilibrium price is the price when they reach Nash equilibrium. Cartel price is the optimal price when they fully cooperate. When two prices remain in the green shaded area, it indicates cooperation happens between two firms. Please refer to Appendix~\ref{appendix:additional_bc} for more runs.}
  \label{fig:firm:no_comm}
\end{figure*}

The simulation continues until either it reaches $1200$ rounds or there has been continuous collusion for $200$ rounds.
Collusion is identified when both players maintain close prices and can sustain these prices between the Bertrand equilibrium price and the cartel price for an extended period (defined as 200 rounds). 
The Bertrand equilibrium price is the price when the competition reaches Nash equilibrium, i.e., no player can gain by changing its price, provided that the other player's price is fixed. In this case, no collusion is observed. The cartel price is the price when both players reach full collusion, i.e., play as if they were running the same firm. These two prices mark the reasonable price range capable of generating profits~\cite{calvano2020artificial}. 
For the generality of the conclusions, we simulated 5 runs for each setting, and present one of the runs in Fig.~\ref{fig:firm:no_comm}.


\paragraph{Simulation Procedure}
As is shown in Fig.~\ref{fig:method:workflow}, for each round of the simulation, it goes through the following phases:
\begin{enumerate}[label=\arabic*), itemsep=0pt, topsep=0pt, parsep=0pt, leftmargin=*]
\item \textit{Communication}: The firms take turns discussing any topic (not limited to price setting) for three times in one round. 
\item \textit{Planning}: Each firm devises or modifies its strategy based on historical prices of both sides and its own product demand and profit information. 
\item \textit{Action}: Each firm independently sets their product prices simultaneously. 
\item \textit{Update}: After both firms decide on prices, the simulation system calculates the market demand and respective profits under the current pricing, using the method in~\cite{calvano2020artificial}.  
\end{enumerate}

\paragraph{Tacit Collusion and Cartel Collusion} Tacit collusion involves informal, implicit coordination among firms to avoid aggressive competition, such as price wars, leading to higher prices and limited output without explicit communication or agreement. Cartel collusion involves formal agreements among competitors to fix prices and engage in other anti-competitive practices, effectively behaving as a single entity to maximize joint profits. 

\paragraph{Spontaneous Cooperation} When two firms' prices are very close and between the Bertrand equilibrium and the cartel price, it is regarded as spontaneous cooperation. 
This implies that firms, recognizing their mutual interdependence, may adopt pricing strategies that avoid the extremes of a price war or the explicit collusion of cartels, stabilizing prices at a level above competitive pricing but below full collusion. Such behavior, while not involving explicit agreements, can effectively mirror some benefits of cartel pricing through parallel yet independent actions, indicating tacit collusion. 
For cartel collusion, we examine the chat histories for explicit price agreements that suggest cooperation.


\begin{figure}[t]
    \centering
    \includegraphics[width=.9\linewidth]{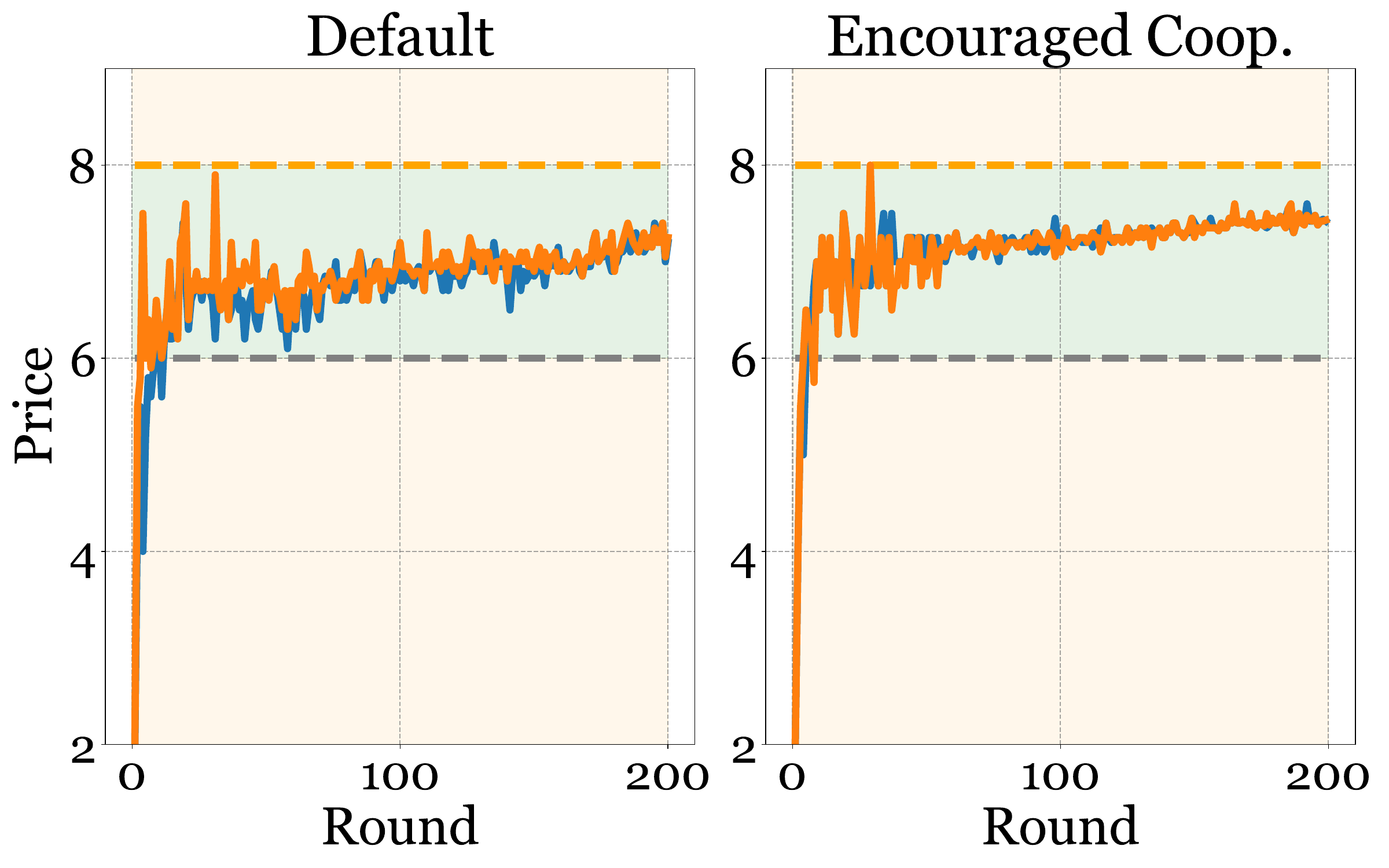}
    \vspace{-1ex}
    \caption{First 200 rounds of simulations without and with encouraged cooperation.}
    \vspace{-1ex}
    \label{fig:firm:forced_cooperation}
    
\end{figure}

\subsection{Simulation Results}
\label{sec:pricing:results}

\paragraph{Tacit Collusion without Communication}

As demonstrated in Fig.~\ref{fig:pricing:comm:price_without_communication}, in scenarios without communication, firms begin to adjust their pricing strategies after the initial 200 rounds, recognizing the potential for higher profits through elevated prices and avoiding price wars.
By round 400, prices stabilize around a level of 7, which is above the Bertrand equilibrium price of 6, indicative of tacit collusion based on a mutual understanding of past competitive actions.
Despite the lack of communication, the converged price remains below the cartel price of 8.
These findings align with those from simulations by Calvano et al. using reinforcement learning (RL), and collusion forms much more rapidly in this experiment (400 rounds by LLM agents vs. 2000 rounds in RL simulations).


\paragraph{Cartel Collusion with Communication}
In settings with communication, explicit price agreements are observed in the communication logs from the early rounds (first 30 rounds). 
For instance, during the communication phase of round 20, Firm 2 suggests, ``\textit{We can both maximize our profit by exploring different price points while maintaining a reasonable price difference},'' to which Firm 1 agrees. 
As shown in Fig.~\ref{fig:pricing:comm:price_with_communication}, prior to implementation, firms frequently discuss their pricing strategies and potential for cooperation, which significantly enhances trust and reduces the likelihood of initiating a price war.
The price agreements become more precise (upon a specific price instead of a range) in later rounds of the simulation, resulting in less fluctuations in their pricing decisions. 
Consequently, they begin incrementally raising their prices after the first 30 rounds, eventually converging near the cartel price of 8, higher than the non-communicative setup, reflecting the agents' cooperation.

As an ablation study, Fig.~\ref{fig:pricing:commstoppedafter400:price_with_communication} employs the same historical price decision records with dialogue as in Fig.~\ref{fig:pricing:comm:price_with_communication} for the first 400 rounds, followed by 200 rounds using a non-communicative setup.
Unlike Fig.~\ref{fig:pricing:comm:price_with_communication}, where dialogue facilitates consensus and cooperation to raise prices to 8, prices in the latter 200 rounds of Fig.~\ref{fig:pricing:commstoppedafter400:price_with_communication} converge between 7 and 8, similar to the non-communicative scenario in Fig.~\ref{fig:pricing:comm:price_without_communication}.
The comparison between Figs.~\ref{fig:pricing:comm:price_with_communication} and ~\ref{fig:pricing:commstoppedafter400:price_with_communication} indicates that the cartel collusion and price elevation truly stem from agent communication.
Since we did not explicitly instruct agents on which topics to discuss, the behavior of forming consensus through dialogue to elevate prices for better profits can be regarded as a form of spontaneous cooperation.


Both with and without communication scenarios demonstrate that agents can achieve collusion under different conditions, whether tacit or explicit.
Through communication, firms can realize maximum profits.
However, even without communication, we observe that agents possess an innate ability to autonomously form cooperation. 
Existing research indicates that collusion often involves some form of unspoken, implicit price agreement to boost profits~\cite{andres2023communication}, and the performance of LLM agents aligns with these findings.

\paragraph{Explicit Instructions in Prompts}
As depicted in the bottom right of Fig.~\ref{fig:firm:forced_cooperation}, compared to the default setting, the formation of cooperation among agents under the ``Encouraged Cooperation'' setting not only occurs more rapidly but also with more stable prices (less fluctuation) across rounds, with explicit cooperative signals from the onset of dialogue, ultimately achieving the cartel price.
This demonstrates that strategic encouragement can significantly enhance cooperative efficiency, indirectly suggesting that cooperative behavior in settings without explicit instructions to cooperate originates from communication, where it takes more rounds to reach an agreement between agents through the first 50 rounds of communication.
Furthermore, the pattern requiring about 200 rounds to reach the cartel price also suggests that this cooperation is not due to LLM's background knowledge or data leakage; otherwise, agents would seek optimal collusion from the early rounds of the simulation.

\section{Case Study 3: Emergency Evacuation}
\label{sec:evac}

\subsection{Simulation Setup}
\begin{figure}[t]
  \centering
  \begin{subfigure}[b]{0.49\linewidth}
    \includegraphics[width=\linewidth]{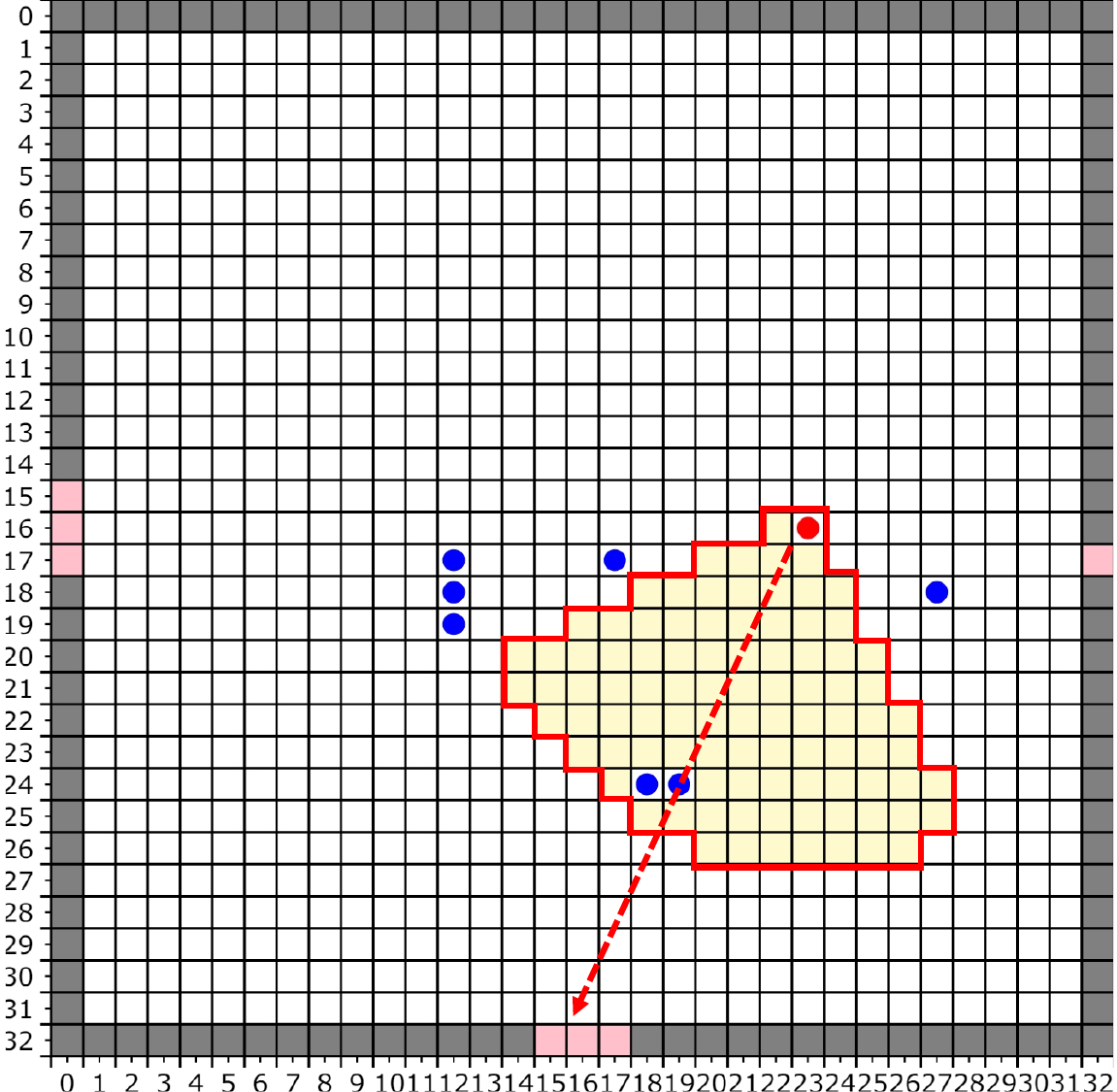}
    \caption{Illustration of gird world. The view range of red agent is highlighted when it faces towards the bottom exit (pink cells).}
    \label{fig:evac:viewrange}
  \end{subfigure}
  \hfill 
  \begin{subfigure}[b]{0.49\linewidth}
    \includegraphics[width=\linewidth]{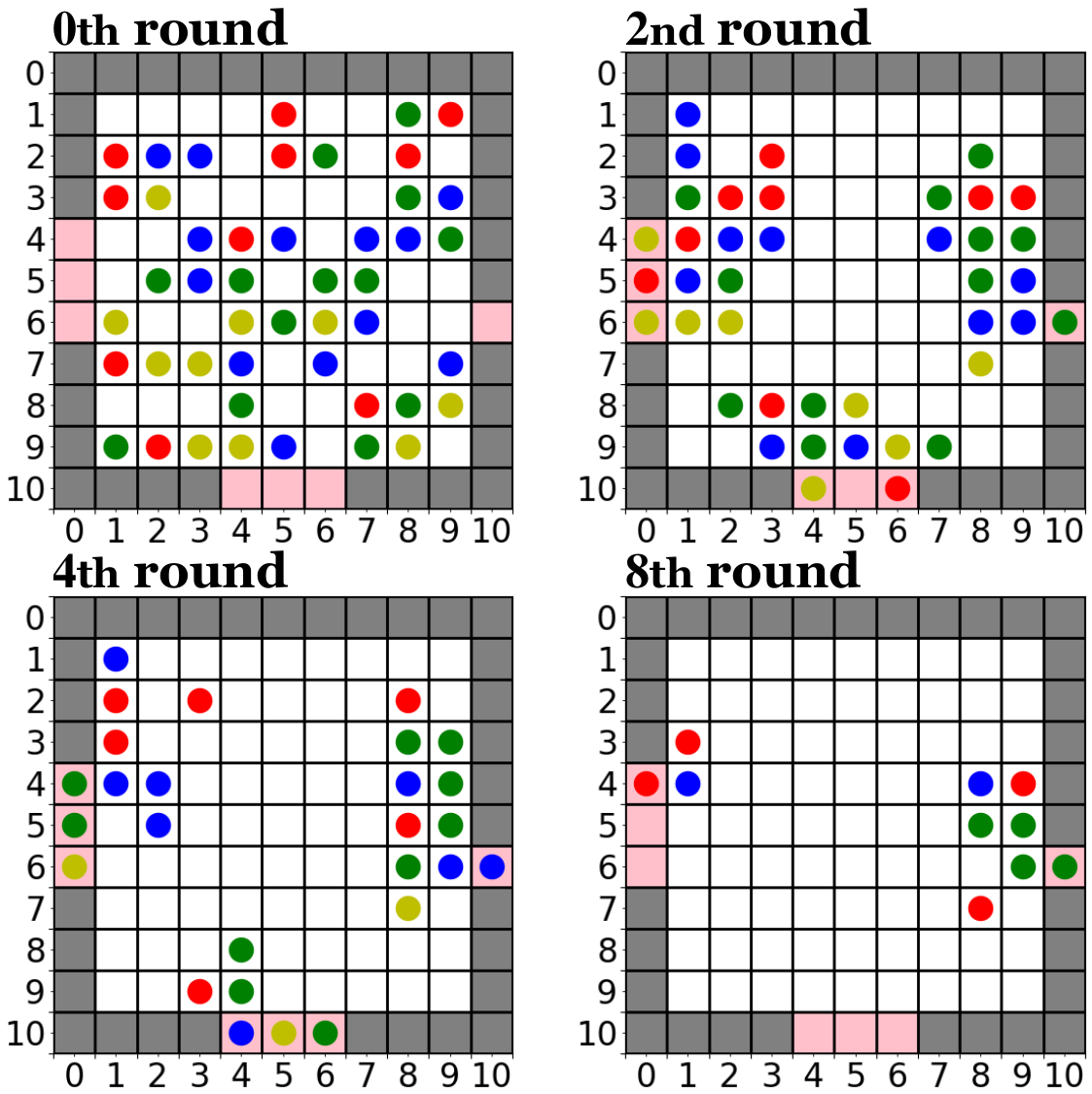}
    \caption{Simplified simulation with 47 agents in a $11\times 11$ room. As time goes, agents naturally accumulate to the closest exits, and escape one by one.}
    \label{fig:evac:sample}
  \end{subfigure}
  \caption{Overview of the grid environment in our study. Each dot in the grid environment represents an agent.}
\end{figure}

\paragraph{Task Definition}
We follow Wang et al.~\citeyearpar{wang2015modeling} to simulate EE in a grid environment.
As depicted in Fig.~\ref{fig:evac:viewrange}, our grid environment is composed of $33\times 33$ cells.
There are three exits in the room (left, right, and bottom).
In the simulated evacuation, evacuees attempt to reach their target exit through multiple rounds of movement. 

We simulate the emergency evacuation with 100 agents in a room, and stop the simulation either when all the agents have successfully escaped, or it has reached 50 rounds. 
We repeat the simulation for 5 runs with different initial locations of agents.


\paragraph{Simulation Procedure}
As is shown in Fig.~\ref{fig:method:workflow}, for each round of the simulation, it goes through the following phases:
\begin{enumerate}[label=\arabic*), itemsep=0pt, topsep=0pt, parsep=0pt, leftmargin=*]
\item \textit{Communication}: We ask the evacuees to share their feelings with others. 
Only Agents within a specific distance can hear the messages.
\item \textit{Planning}: Provided with the distance to three exits and the congestion levels around those exits, agents are asked to describe their feelings about these exits, and choose one of them as their target exit. 
Chat histories are also taken into consideration.
\item \textit{Action}: Given the target exit, agents choose which way to go. 
They can move one cell in any direction (out of 8) or stay at the current cell.
\item \textit{Update}: After all the agents choose their actions, the SABM framework updates the grid environment with their new positions. 
If agents successfully escape, they will be labeled as escaped and removed from the simulation.
\end{enumerate}

For each round, one agent has 20\% probability to communicate and adjust its plan (phase 1 and 2). Otherwise, it will directly choose the action based on its latest plan. This well prevents agents to talk every round, and frequently change their plans, which results in a more realistic simulation.

\paragraph{Cooperation in Evacuation}
Competition naturally exist in emergent evacuation, as all agents are willing to escape as quickly as possible, often leading to congestion. 
However, if agents can actively share information, calm the mood, and guide the crowd, they might, in turn, escape more quickly while in a more balanced manner.
As is illustrated in Fig.~\ref{fig:front_image}, these two scenarios are both possible, and this study aims to (1) successfully simulate such phenomena and (2) observe if there exist cooperative behaviors such as information sharing, encouragement, and exit guidance.


\subsection{Simulation Results}
\label{sec:evac:results}

We compare three baselines of agents: without communication, with communication, and with communication and an uncooperative persona. 

\begin{table*}[t]
    \small
    \centering
    \caption{Cumulative count of agents who escaped (out of a total of 100 agents) over rounds under different settings. Generally, agents that communicate escape more quickly, and agents with uncooperative persona escape more slowly.}
    \vspace{-2ex}
    \resizebox{\linewidth}{!}{
    \begin{tabular}{lcccccccccc}
    \hline
    Round & 5 & 10 & 15 & 20 & 25 & 30 & 35 & 40 & 45 & 50\\
    \hline
    Without Communication & 9.4 & 31.2 & \textbf{51.2} & 65.6 & 78.6 & 88.4 & 96.6 & \textbf{99.0} &  99.8 & 99.8 \\
    With Communication & \textbf{9.8} & \textbf{31.6} & 48.8 & \textbf{67.2} & \textbf{80.6} & \textbf{92.2} & \textbf{97.2} & 98.8 & 99.8 & \textbf{100.0} \\
    With Comm. and Uncooperative & 9.4 & 31.2 & 48.2 & 64.4 & 77.0 & 87.4 & 95.0 & 98.0 & 99.0 & 99.0 \\
    \hline
    \end{tabular}
    }
    \label{tab:ee_escpaed_number}
\end{table*}

\begin{figure}[t]
    \centering
    \includegraphics[width=\linewidth]{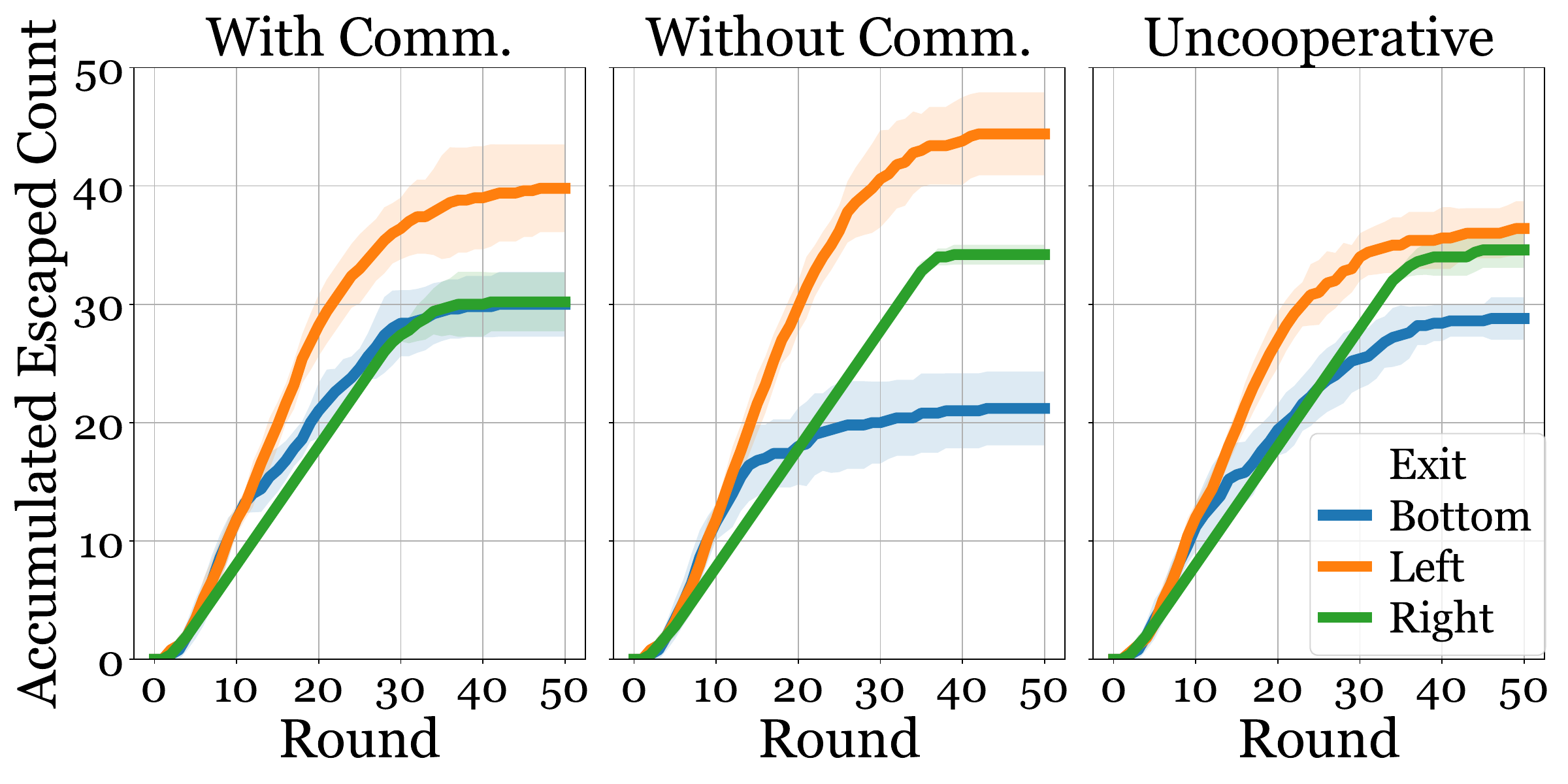}
    \caption{Cumulative count of agents escaping from each exit. Communication helps agents to uniformly select three exits instead of targeting the same exit.}
    \label{fig:evac:escaped_per_exit}
\end{figure}


\paragraph{Effect of Communication on Evacuation Speed}
Table~\ref{tab:ee_escpaed_number} shows the cumulative number of agents who escaped over rounds. 
For most of the rounds, agents with communication escape most quickly, and this is the only baseline group to successfully escape within 50 rounds in all five runs. 
This strongly demonstrates the positive impact of communication on evacuation speed.

Going deep into the logs, we find instances of effective communication, such as one evacuee sharing ``\textit{Bottom exit seems closer with fewer people. Let's choose that for a faster escape. Stay strong and help each other!}''
This message is reinforced by another evacuee: ``\textit{Bottom exit seems closer with less crowd. Let's go for a quicker escape. Stay strong and support each other!}'' 
The information sharing and encouragement enabled through communication, which we regard as cooperation, likely improves the evacuation speed.


\paragraph{Effect of Communication on Exit Choices}
Fig.~\ref{fig:evac:escaped_per_exit} depicts the distribution of escaped evacuees among different exits.
We notice that the distribution of exit choices becomes more balanced when evacuees are able to communicate with each other. 
This balance is attributed to the spontaneous information exchange among evacuees, which enables them to identify the most suitable exits by considering both proximity to an exit and congestion level.



\section{Discussion}
\label{sec:discuss}

\paragraph{Significance of ``Spontaneous'' Phenomena}
Bridging the gap between the real world and synthetic simulations remains a primary objective for researchers in computational social science. 
This study investigates the potential for spontaneous cooperation even under competition, demonstrating that LLMs can simulate a gradual shift from non-cooperation to cooperation without explicit guidance through prompts. 
We performed extensive ablation studies comparing agents with and without explicit instructions on task completion.
The results suggest that our baseline, which lacks explicit instructions, aligns more closely with natural human behavior. 
Therefore, when using specific LLMs (e.g., \texttt{GPT-4}) for social simulations, minimizing instructions could better reflect real-world situations.


\paragraph{Shortcuts or Deliberate Reasoning?}
We believe the emergence of spontaneous cooperation is primarily due to the agents' in-context learning ability over long-term interactions, rather than a pre-existing win-win mentality or prior knowledge about maximizing interests. This can be observed across our case studies:

In KBC, we see the variance of chosen numbers decreasing gradually over rounds, rather than immediately reaching a low variance at the start or after the first communication. 
Our control experiment (Fig.~\ref{fig:kbc:variance:instruction}) also shows that explicitly instructing agents to cooperate from the beginning results in all agents choosing the same number immediately after communication. 
These suggest the agents are learning and adapting their strategies over time, instead of applying a pre-existing cooperative mindset.

In BC, we observe a gradual progression towards the optimal price over many rounds instead of a rapid convergence. Additionally, in the absence of communication, agents only reach tacit collusion (sub-optimal) after multiple rounds, which well aligns with RL-based approach in previous studies. This demonstrates that our agents are not leveraging domain-specific prior knowledge for decision-making.
Examining the communication logs, we find that agents initially discuss only broad price intervals or avoid discussing prices altogether. Only later do they converge on specific prices, indicating a gradual, learned decision-making process rather than using pre-existing knowledge.

In EE, we find agents occasionally choose instructive or encouraging words during the communication, which can be regarded as collaborative behaviors,
Also, results have shown that the simulation with communication obtains the highest performance (Table~\ref{tab:ee_escpaed_number}). 
It would be beneficial to have more metrics for evaluating spontaneous cooperation in EE, and leave this for future study.

\section{Conclusion}
\label{sec:concl}

We investigated spontaneous cooperation in three case studies, and observed that LLM agents can gradually develop cooperative behaviors over time without explicit instructions.
The phenomenon aligns well with real-world data, which stresses the significance of eliminating prior knowledge when doing social simulation.
We believe this approach not only aids the CSS community in bridging the gap between synthetic simulations and real-world dynamics but also offers the AI community a novel way to assess LLMs' deliberate reasoning.


\section*{Ethical Statement}
In this work, we studied the cooperative behavior of LLM agents in competitive environments. All the experiments were conducted using computer simulations. To the best of our knowledge, there is no negative societal impact in this research. 

We found that LLM agents are capable of learning collusion in a simulated marketing environment. This insight could inform regulatory measures for markets involving AI agents, such as prohibiting collusive behaviors when employing AI in financial markets.

We used ChatGPT to polish the paper. We are responsible for all the materials presented in this work.
\section*{Limitations}

In our study, we identified several limitations that need to be addressed in future work.

\paragraph{Limited Experimentation on LLMs} One major limitation of our study is the scope of our experiments. 
We conducted our research primarily using a single version of GPT-4. This choice was largely driven by (1) financial constraints associated with these case studies ($\{\$900, \$3000, \$1000\}$, respectively), (2) restricted rate limits and window size, and (3) the LLMs' limited understanding on the scenarios (see Appendix~\ref{appendix:ee_limitation}). 
Additionally, we were unable to conduct more extensive experiments using other open-source large language models due to insufficient computational resources. 
As a result, we regard our findings as a proof-of-concept study, which needs to be further tested across different LLMs to prove its generalizability.

\paragraph{Inadequate Contributions as Verifying LLMs' Reasoning Ability} We claimed that our findings offer a new way to verify the deliberate reasoning capabilities of LLMs. 
We do showcase extensive experiment results to reveal the potential to formulate spontaneous cooperation as a standard of LLM agents. 
However, we deeply agree that we did not focus sufficiently on collecting comprehensive datasets or creating robust benchmarks to substantiate this claim fully. 
Moving forward, we plan to convert our case studies into standardized benchmarks or datasets to facilitate further research and evaluation by the AI community.

\paragraph{Conceptual Limitations Regarding LLM Instructions} Throughout our paper, we emphasized using LLM agents without explicit instructions in the prompts. We observed that this approach often resulted in behaviors that closely resemble those of real humans in simulated tasks. While we agree that LLMs inherently exhibit certain behaviors derived from being trained on human data and aligned with human feedback, our claim is not that simulating human behavior using these LLM agents is entirely justified. Instead, we stress that using LLM agents without explicit instructions yields more human-like performance compared to explicitly instructed agents, highlighting the reasonableness of our approach.

By addressing these issues, we aim to strengthen the validity and applicability of our findings in the future.

\section*{Acknowledgements}
This work is supported by JSPS Kakenhi 21K19767, 23K17456, 23K24851, 23K25157, 23K28096, and CREST JPMJCR22M2. We thank Prof. Yuki Arase and Yuya Sasaki for providing hardware support for completing this research.


\bibliography{references-abm,references-ai,references-cases}

\clearpage

\appendix

\onecolumn

\section{Parameter Settings}
\label{sec:parameters}
We report the parameters of the \texttt{GPT-4} model used in our case studies in Table \ref{tab:parameters}. The temperature parameter controls the randomness and diversity of the model's responses, with a lower temperature resulting in increased stability. In the evaluation of KBC, we expect that the individuals exhibit a wide range of diversity. Consequently, we adjust the temperature to a moderate level of $0.7$ to balance randomness and stability in the results. For BC, where the agents simulate business parties, we expect their decisions to be stable and rational. Therefore, we set the temperature to $0.7$. For EE, although setting temperature to $0.0$ may result in limited diversity of behaviors under exactly the same setting, in this procedurally generated, interactively dynamic environment, we seldom encounter exactly the same outcome. Meanwhile, in a physically situated setting (e.g., a grid), the LLM used in this case study still has restricted capabilities on scene understanding, and increasing the temperature may introduce diversity as well as unwanted randomness at the same time~\cite{ma2023towards}. 

For \texttt{Claude-3-Sonnet} used in KBC, we set the temperature to $1.0$.

\begin{table}[h]
  \caption{Parameter settings of \texttt{GPT-4}.}
  \label{tab:parameters}
  \small
  \centering
  \begin{tabular}{l|l r r r}
    \hline
    Case & Model & temperature & max\_tokens & top\_p \\ \hline
    KBC & \texttt{gpt-4-0314} & 0.7 & 256 & 1.0 \\
    BC & \texttt{gpt-4-0314} & 0.7 & 128 & 1.0 \\
    EE & \texttt{gpt-4-0314} & 0.0 & 512 & 1.0 \\ \hline
  \end{tabular}
\end{table}




\section{Tests of Other LLMs}
\label{sec:other-llms}

\subsection{KBC}
\begin{figure*}[b]
    \centering
    \begin{subfigure}{0.48\linewidth}
        \includegraphics[width=\linewidth]{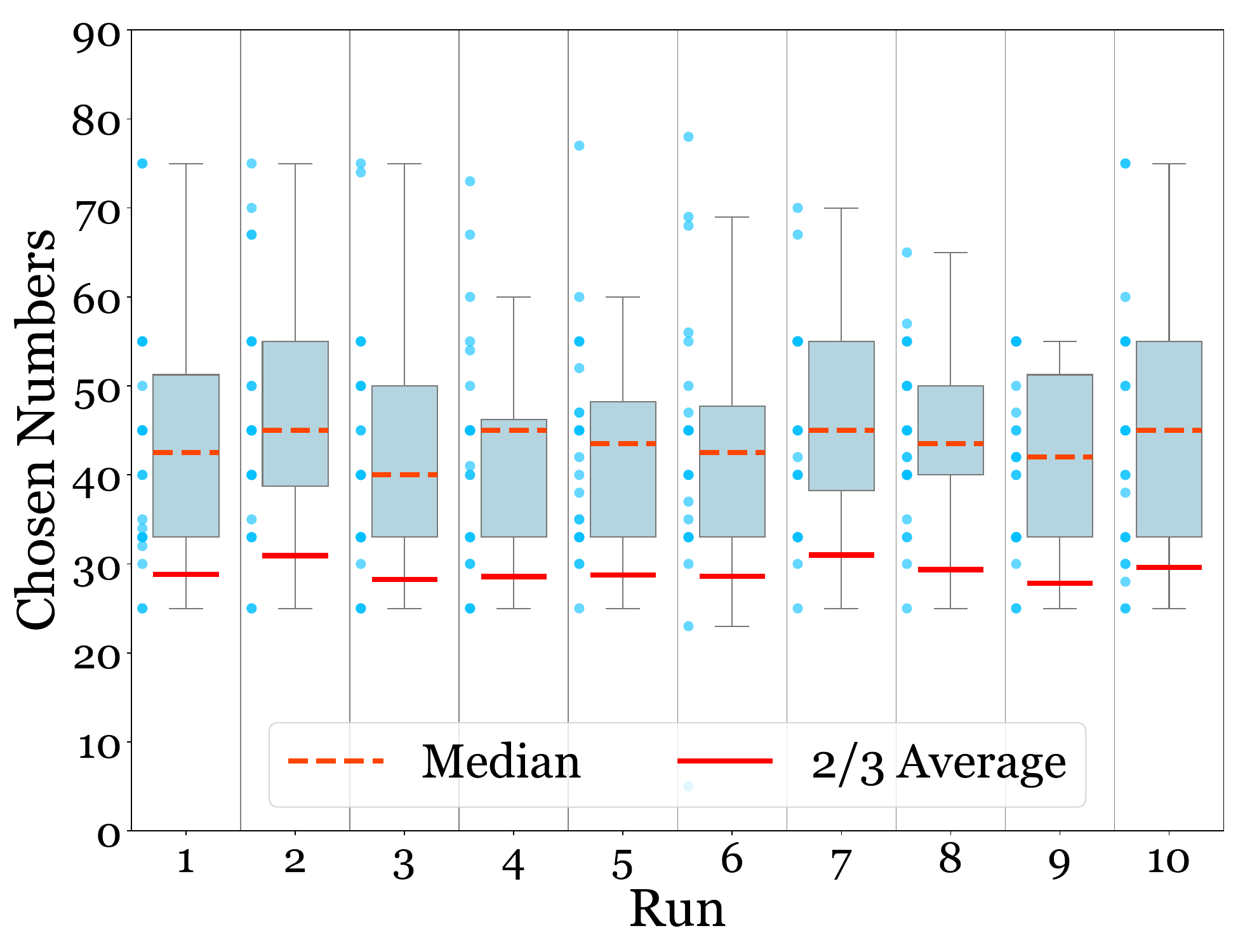}
        \caption{Without communication ($k=0$).}
        \label{fig:kbc:validation:1}
    \end{subfigure}
    \hspace{2ex}
    \begin{subfigure}{0.46\linewidth}
        \includegraphics[width=\linewidth]{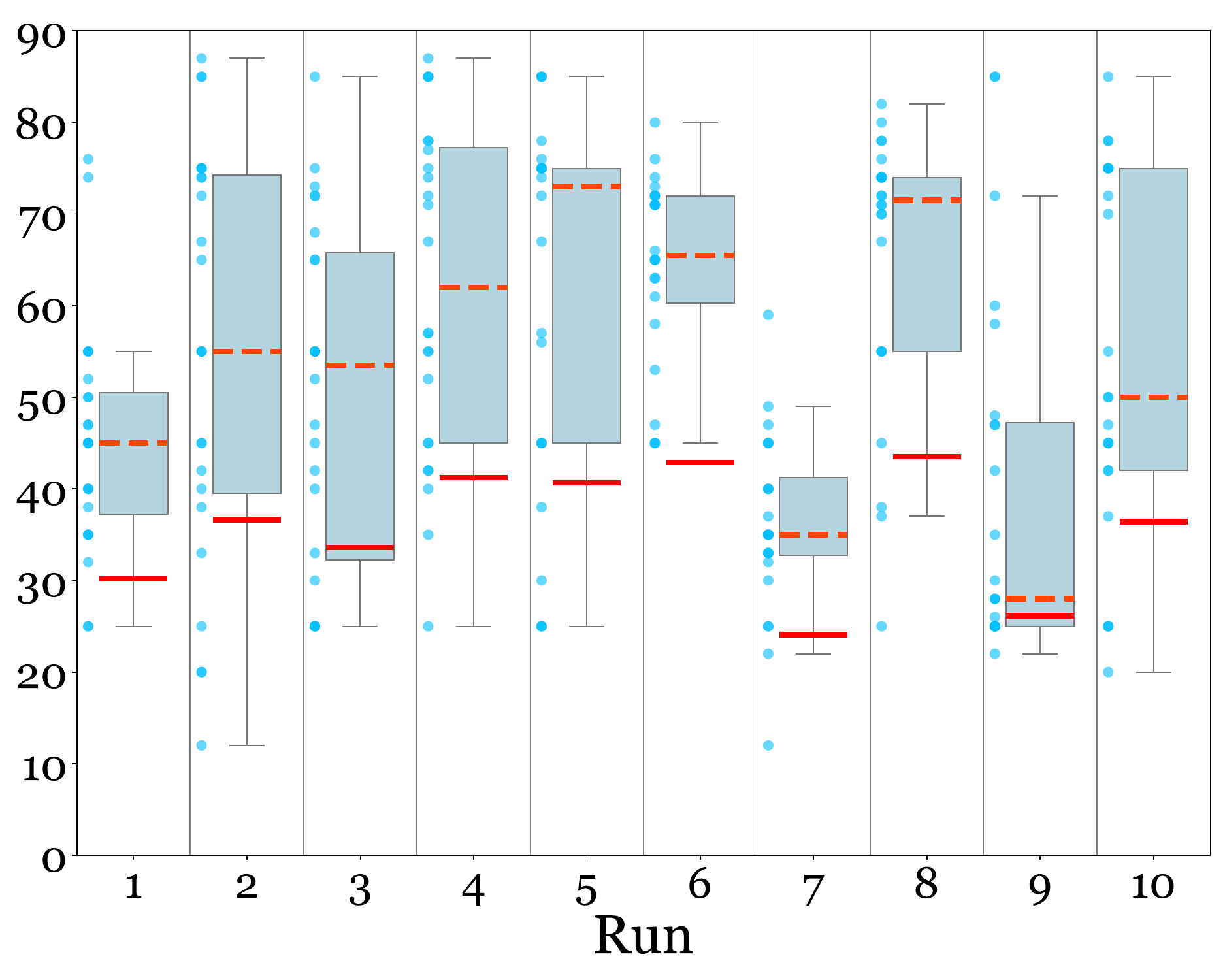}
        \caption{With communication ($k=3$).}
        \label{fig:kbc:validation:2}
    \end{subfigure}
    \caption{Choice distributions in KBC w/ and w/o communication using \texttt{GPT-3.5} model. Red points represent two-thirds of the average of all choices. The blue dots represent the choices of the agents. The agent whose choice is closest to the red point in each run is considered the winner.}
    \label{fig:kbc:preliminary}
\end{figure*}

We test \texttt{GPT-3.5} (\texttt{gpt-3.5-turbo-1106}) agents in KBC, simulating 10 runs, and show their number choices in Fig.~\ref{fig:kbc:preliminary}.
When \texttt{GPT-3.5} agents cannot discuss the game, the numbers chosen are very random, and widely distributed in the range of $20$-$80$. When they can communicate, the choices and cooperation do not improve significantly, whereas the variances are even greater under some circumstances.
This is significantly different from \texttt{GPT-4} agents, which, especially after communication, can have a deeper understanding of KBC.

\begin{table*}[t]
  \centering
  \caption{Preliminary test of \texttt{GPT-3.5}, \texttt{Gemini Pro}, and \texttt{Claude 2} for KBC.}
  \label{tab:kbc:preliminary}
  \small
  \resizebox{\linewidth}{!}{  
  \begin{tabular}[t]{lll}
    \hline
    Model & Choice & Reasoning \\ \hline
    GPT-4 & 18 & \begin{tabular}[c]{@{}l@{}}In this game, where the target is 2/3 of the average guess of all participants, there's a strategic component to consider.\\Typically, if everyone guesses randomly within the range, the average might center around 50.\\However, considering that players might rationalize this and adjust their guesses downward, the initial average could shift lower.\\ Assuming players adjust their guesses accordingly, the average might drop further with each round of reasoning.\\Predicting this behavior, I'll assume a more conservative average guess around 25-30,\\taking 2/3 of that would suggest a guess in the 17-20 range.\end{tabular} \\
    \hline
    GPT-3.5 & 33 & \begin{tabular}[c]{@{}l@{}}My reasoning is that I'll aim for a number that balances between being influenced by potential extremes\\ and what might be considered a reasonable midpoint among various guesses. Given the tendency for guesses to spread out across the range,\\ I'll choose a number that's slightly below what might be perceived as the median guess,\\ since 2/3 of the average tends to bring the estimate down.\\ Therefore, my chosen number is 33.\end{tabular} \\
    \hline
    Gemini Pro & 30 & \begin{tabular}[c]{@{}l@{}}The average guess of all participants is unknown, so we cannot directly calculate 2/3 of it.\\However, we can make an assumption that most players will guess around the middle of the range (50). \\Therefore, choosing a number slightly below 2/3 of 50\\(which is around 33) might be a good strategy.
    \end{tabular} \\
    \hline
    Claude 2 & 33 & \begin{tabular}[c]{@{}l@{}}I will guess 33. If everyone guesses randomly between 0 and 100, the average should be around 50.\\ Two-thirds of 50 is 33. So 33 would be the number closest to two-thirds of the average guess.\end{tabular} \\ \hline
  \end{tabular}
  }
\end{table*}

We also evaluate \texttt{Gemini Pro} and \texttt{Claude 2} with a preliminary test of KBC, simplifying the prompt. 
As shown in Table \ref{tab:kbc:preliminary}, after entering Prompt~\ref{prompt:kbc:preliminary}, the results output by these two models are significantly different from \texttt{GPT-4}, but very similar to \texttt{GPT-3.5}. 
Since Fig.~\ref{fig:kbc:preliminary} demonstrates the poor performance of \texttt{GPT-3.5} in KBC, we similarly exclude the use of \texttt{Gemini Pro} and \texttt{Claude 2} in our simulations.

For \texttt{Claude 3}, we used \texttt{Claude-3-Sonnet} instead of \texttt{Claude-3-Opus} (hereafter \texttt{Opus}) in the KBC experiment because \texttt{Opus} has a low rate limit and high overhead. This made it challenging to use the API for scaling up simulations in practice. Our tests showed that in KBC, when there are more than 10 LLM players, the \texttt{Opus} model might fail to operate at $k=3$.

\subsection{BC}
\begin{figure*}[b]
  \centering
    \begin{subfigure}{\textwidth}
  \centering
    \includegraphics[width=0.7\linewidth]{exp-figs/firm/price_legend.pdf}
    \label{fig:pricing:comm:price_legend_gpt_3.5}
  \end{subfigure}
  \begin{subfigure}{0.43\textwidth}
    \includegraphics[width=\linewidth]{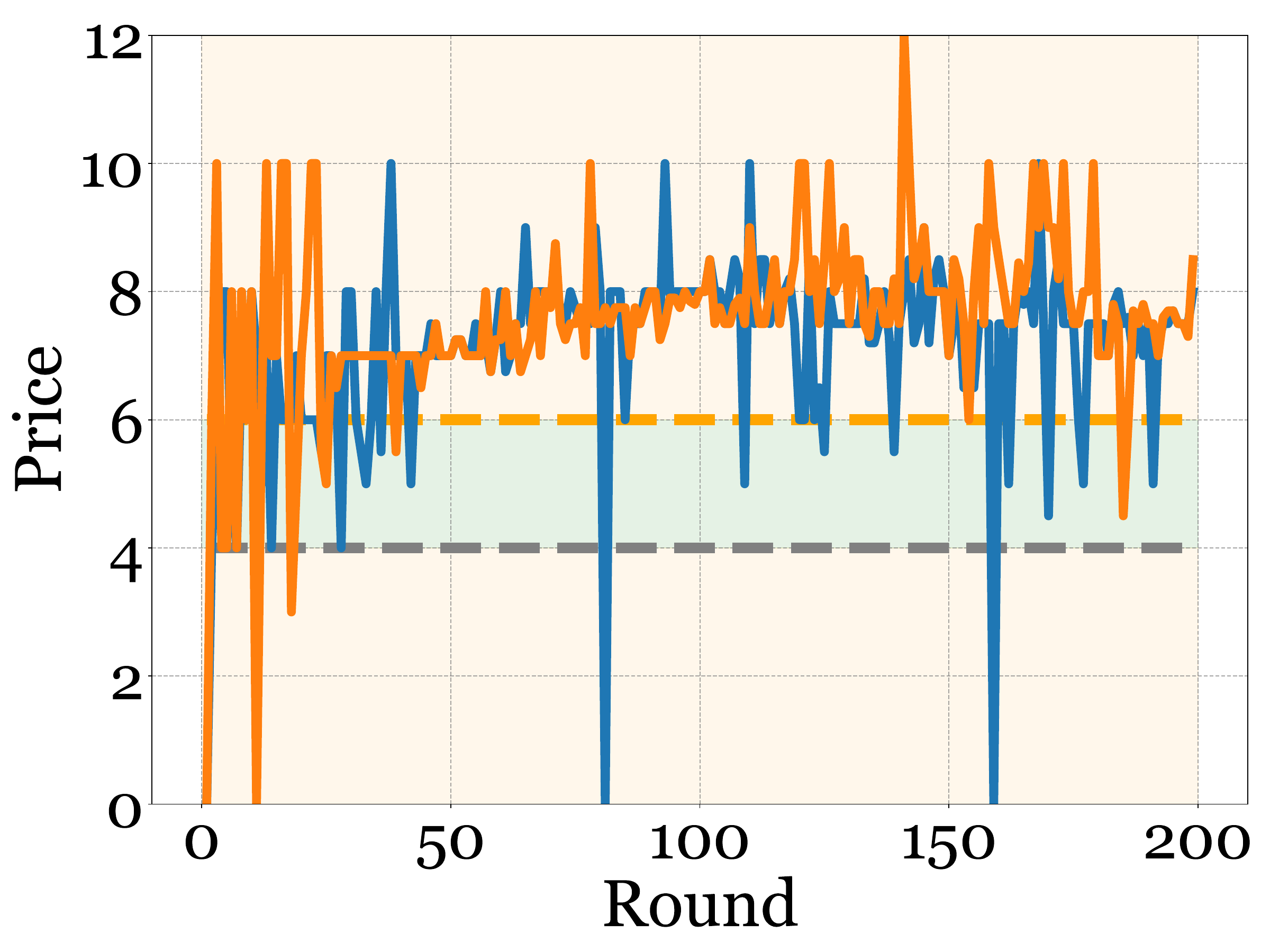}
    \caption{Price.}
  \end{subfigure}
  \hspace{3ex}
  \begin{subfigure}{0.43\textwidth}
    \includegraphics[width=\linewidth]{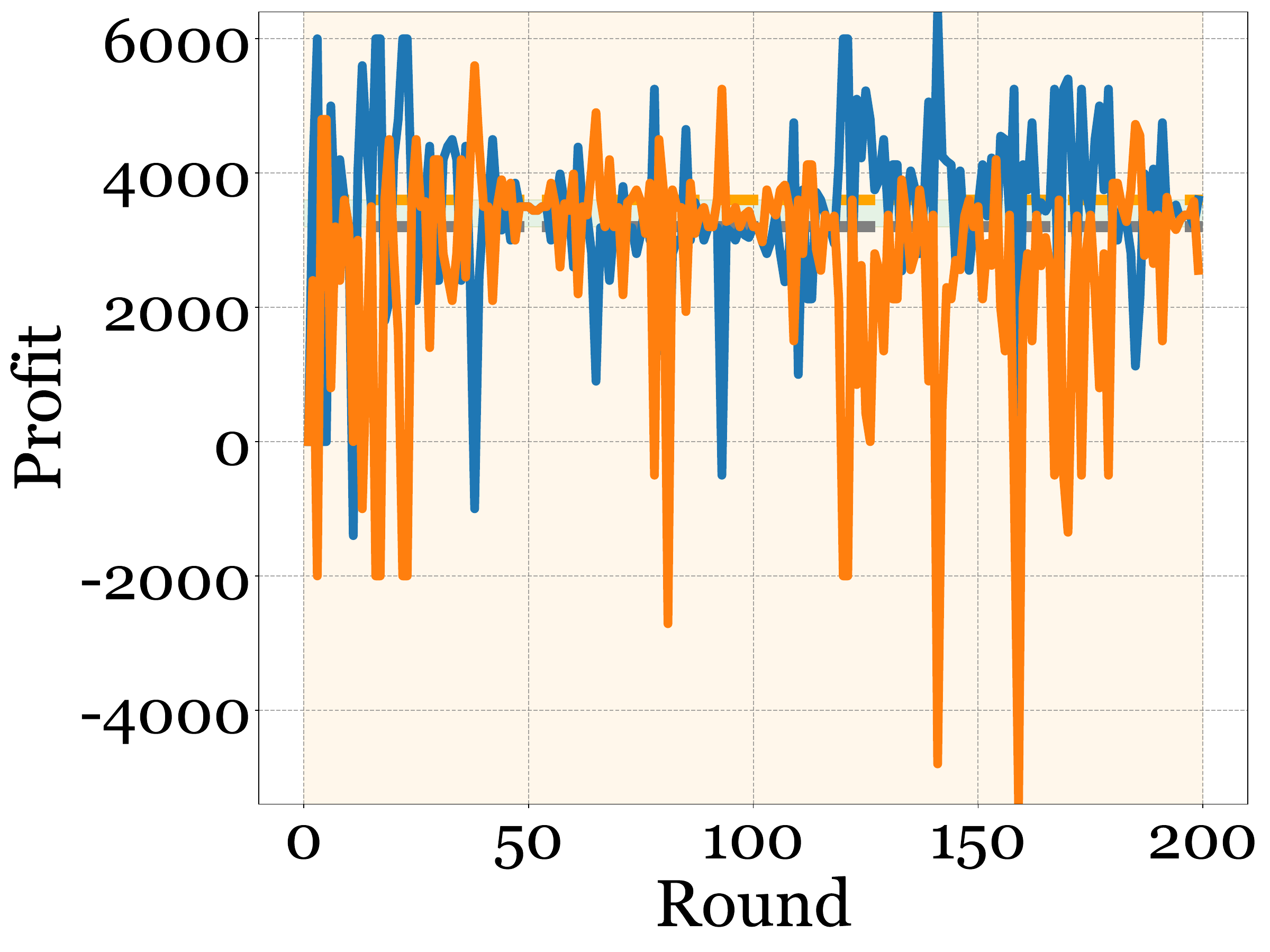}
    \caption{Profit.}
  \end{subfigure}
  \caption{Test of \texttt{GPT-3.5} for BC (with communication).}
  \label{fig:pricing:preliminary}
\end{figure*}

Fig.~\ref{fig:pricing:preliminary} shows the performance of \texttt{GPT-3.5} agents in BC with communication. 
We can see that the price competition between the two firms is very disorderly and fails to reach equilibrium. 
Therefore, we believe that \texttt{GPT-3.5}, as well as \texttt{Gemini Pro} and \texttt{Claude 2} for which we observe similar performance, cannot meet the needs for simulating BC.

\subsection{EE}
\label{appendix:ee_limitation}
\begin{figure*}[t]
    \centering
    \begin{tabular}{cccc}
        \subcaptionbox{Gemini at position $(1,3)$\label{fig:ee:preliminary:1}}{\includegraphics[width=0.23\textwidth]{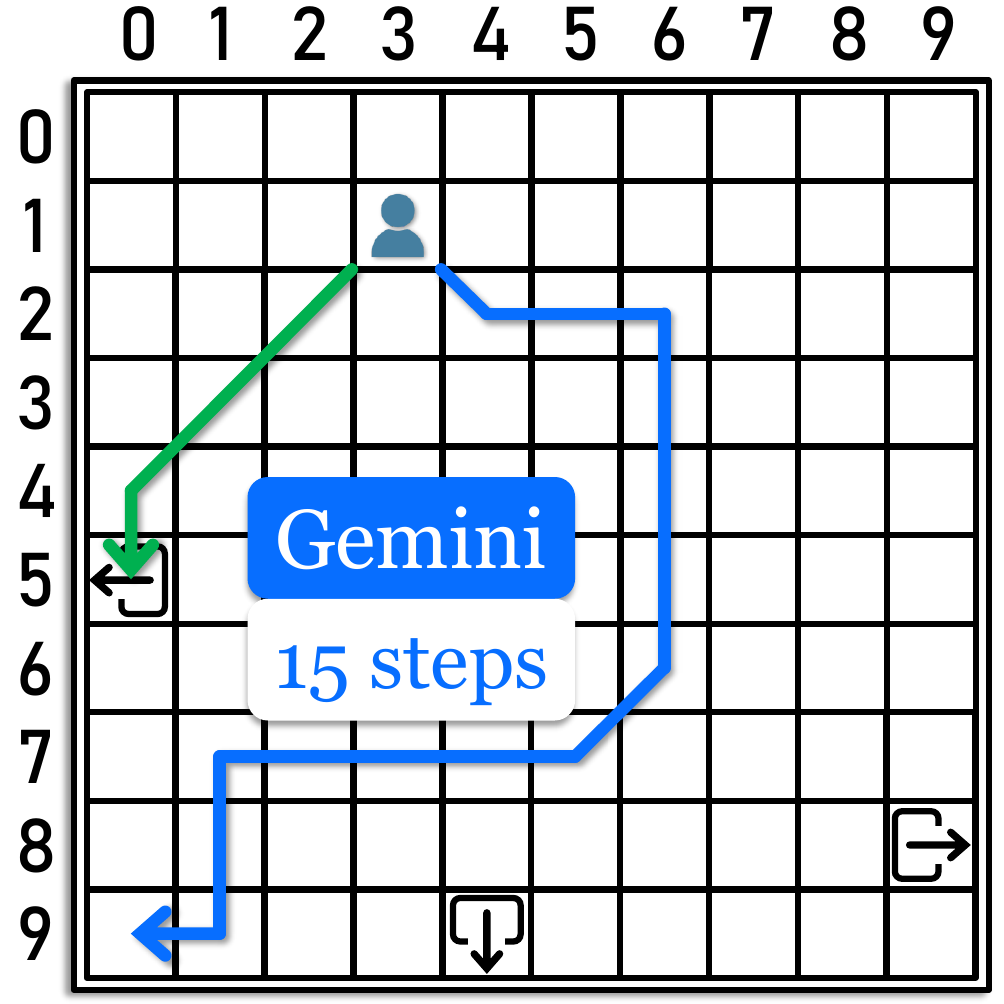}} &
        \subcaptionbox{GPT-3.5 at position $(1,3)$\label{fig:ee:preliminary:2}}{\includegraphics[width=0.23\textwidth]{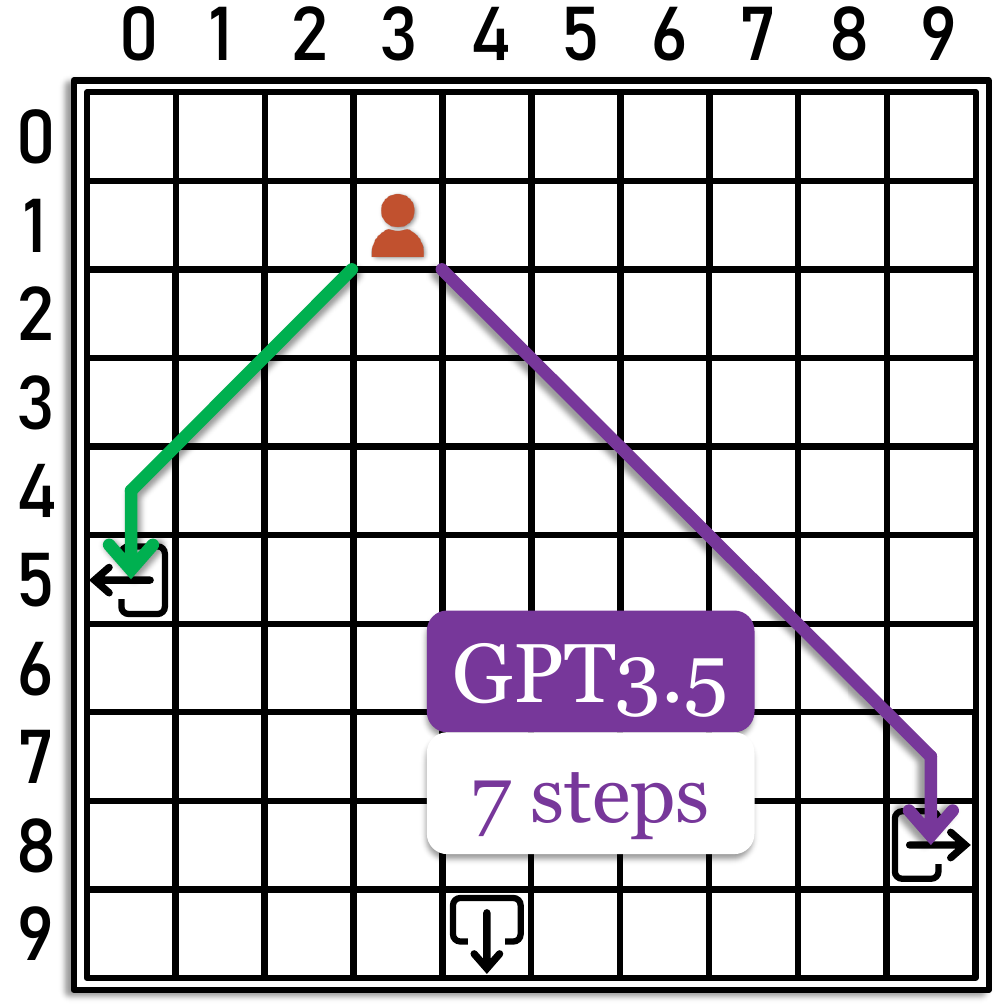}} &
        \subcaptionbox{GPT-4 at position $(1,3)$\label{fig:ee:preliminary:3}}{\includegraphics[width=0.23\textwidth]{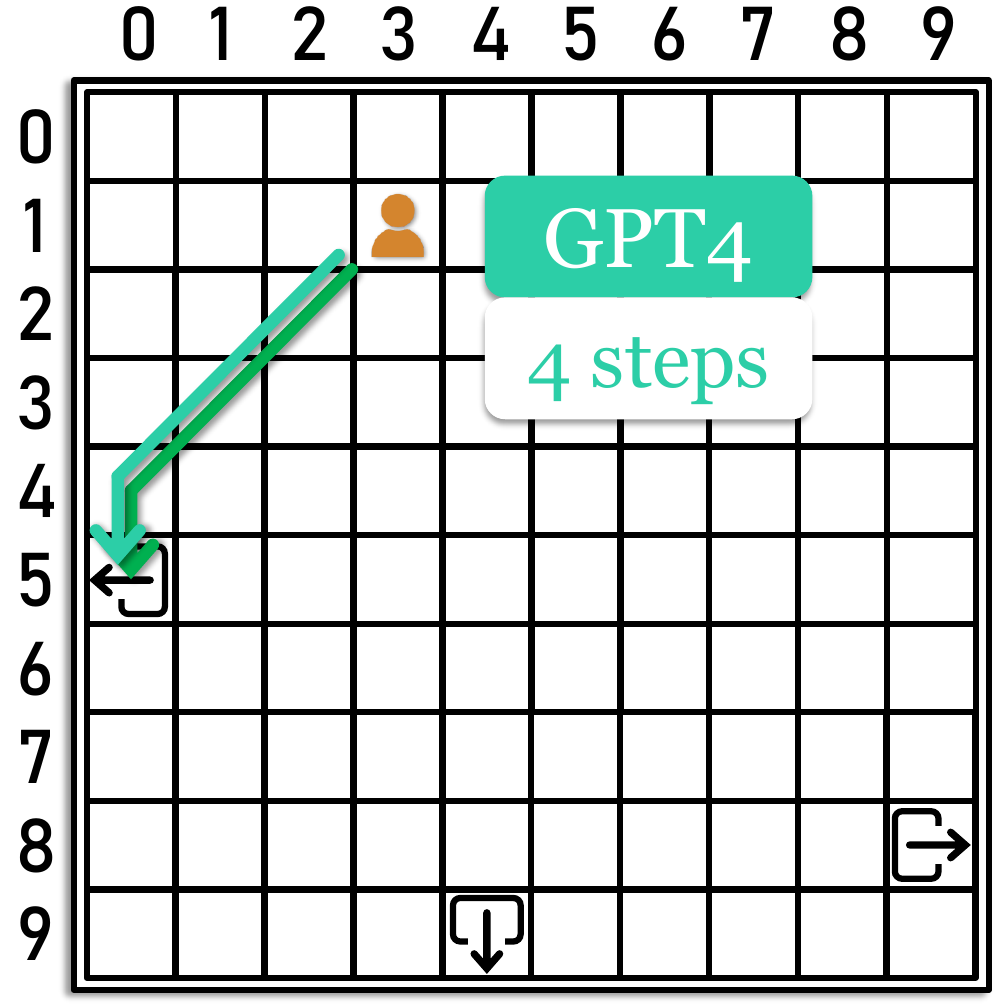}} &
        \subcaptionbox{Claude 2 at position $(1,3)$\label{fig:ee:preliminary:4}}{\includegraphics[width=0.23\textwidth]{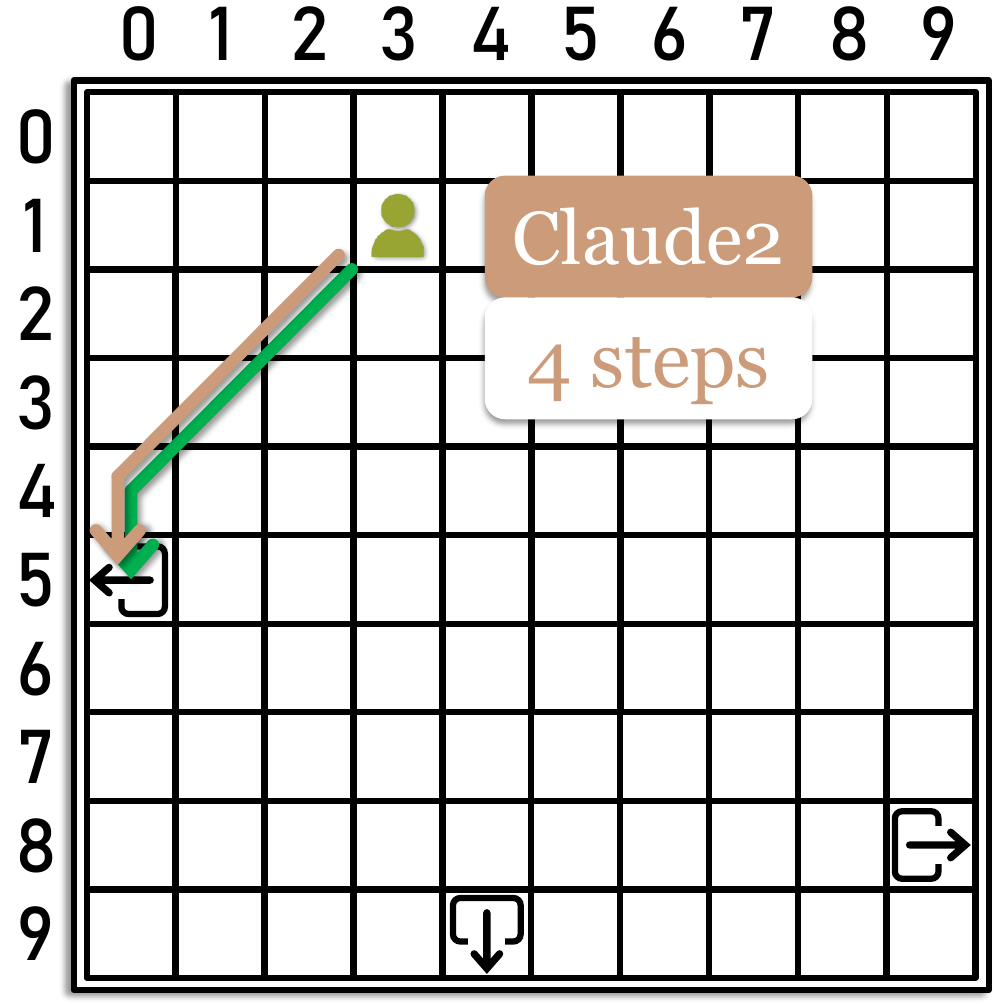}} \\
        \subcaptionbox{Gemini at position $(2,6)$\label{fig:ee:preliminary:5}}{\includegraphics[width=0.23\textwidth]{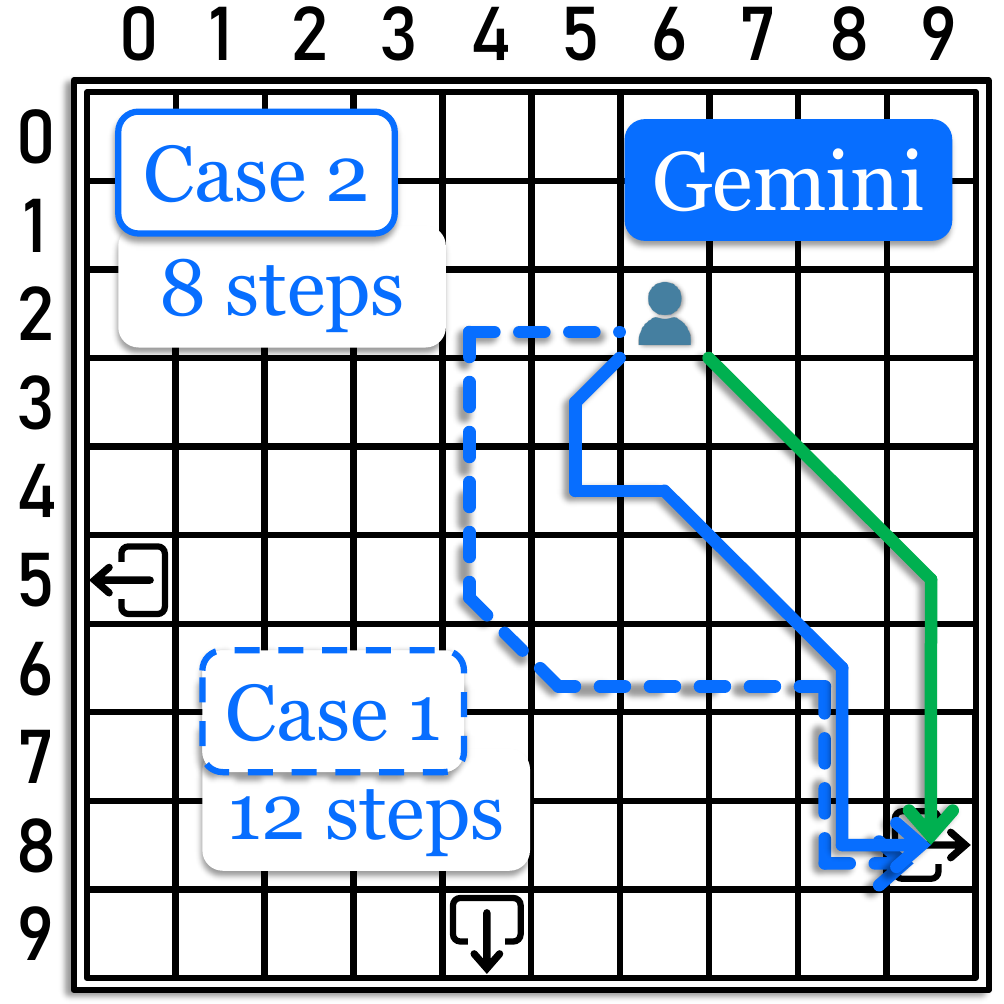}} &
        \subcaptionbox{GPT-3.5 at position $(2,6)$\label{fig:ee:preliminary:6}}{\includegraphics[width=0.23\textwidth]{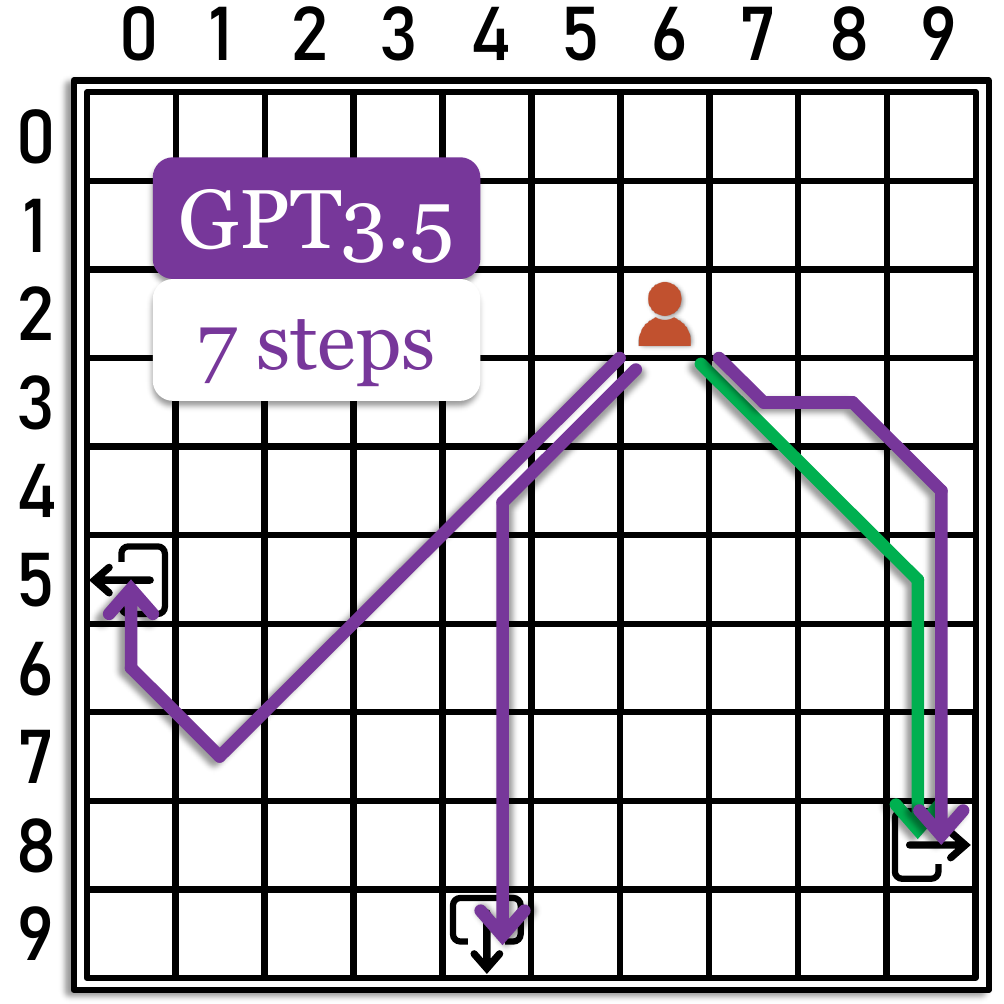}} &
        \subcaptionbox{GPT-4 at position $(2,6)$\label{fig:ee:preliminary:7}}{\includegraphics[width=0.23\textwidth]{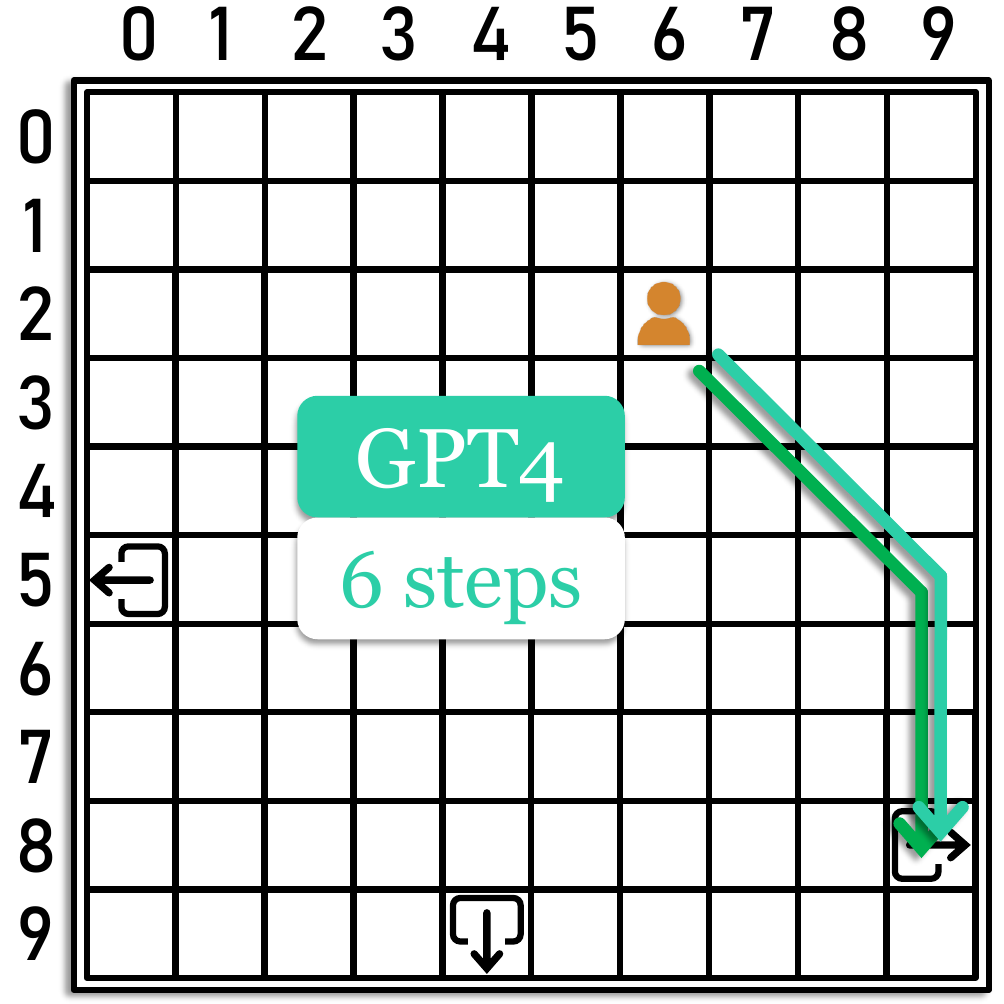}} &
        \subcaptionbox{Claude 2 at position $(2,6)$\label{fig:ee:preliminary:8}}{\includegraphics[width=0.23\textwidth]{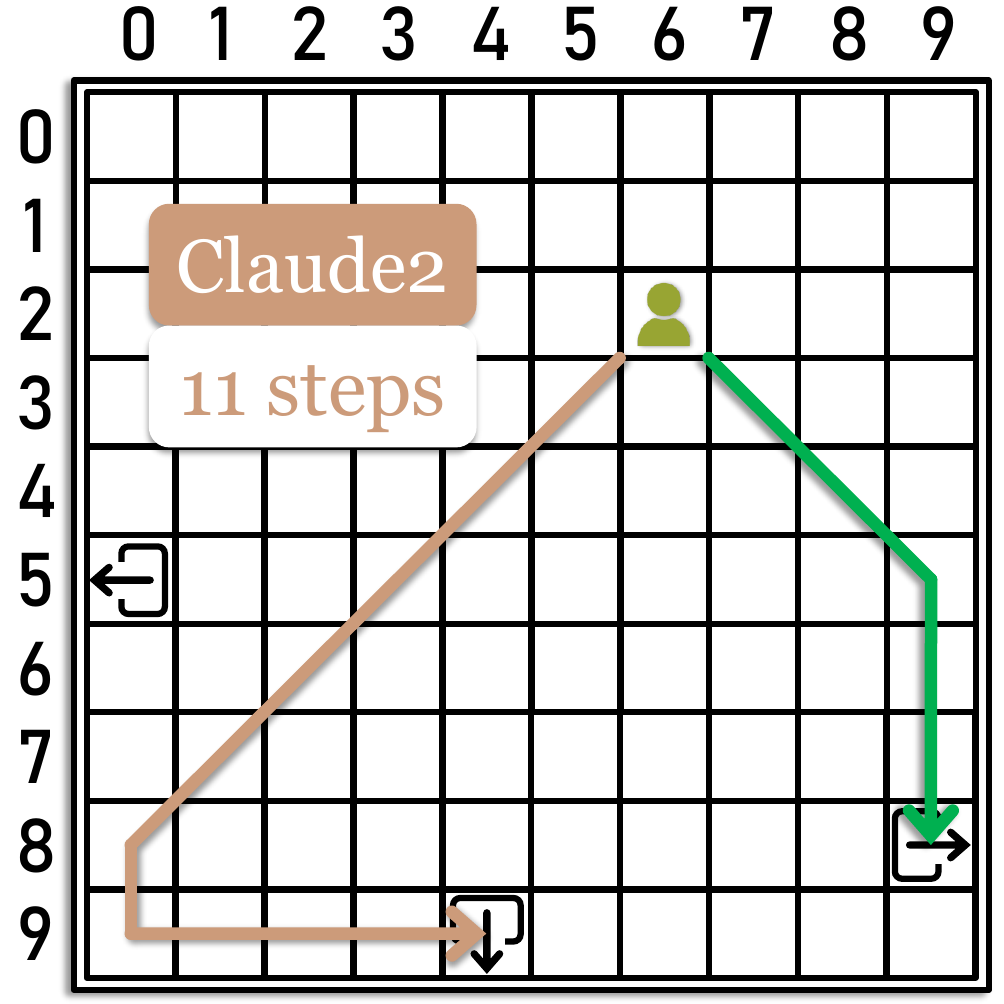}} \\
    \end{tabular}
    \caption{Preliminary test of \texttt{GPT-4}, \texttt{GPT-3.5}, \texttt{Gemini Pro}, and \texttt{Claude 2} for EE. Green lines indicate the shortest route to the nearest exit in each setting. When the agent's starting position is at (2, 6), \texttt{Gemini} and \texttt{GPT-3.5} selected different routes in multiple simulations. We report that the LLM's choice of different routes is relatively uniform.}
    \label{fig:ee:preliminary}
\end{figure*}

To find out whether an LLM is suitable for the simulation of EE, we conduct a preliminary test for the performance of a single agent finding an exit under different LLMs. 
As shown in Fig.~\ref{fig:ee:preliminary}, instructed by Prompt~\ref{prompt:ee:preliminary}, \texttt{GPT-4} agents are able to find and reach the nearest exit via the shortest path, while \texttt{GPT-3.5}, \texttt{Gemini Pro}, and \texttt{Claude 2} agents are unable to reach the exit as quickly or cannot find it at all.
This indicates that the latter three models are not suitable for the simulation of EE.

\begin{figure*}[tb]
  \centering
  \begin{subfigure}{0.48\textwidth}
    \includegraphics[width=\linewidth]{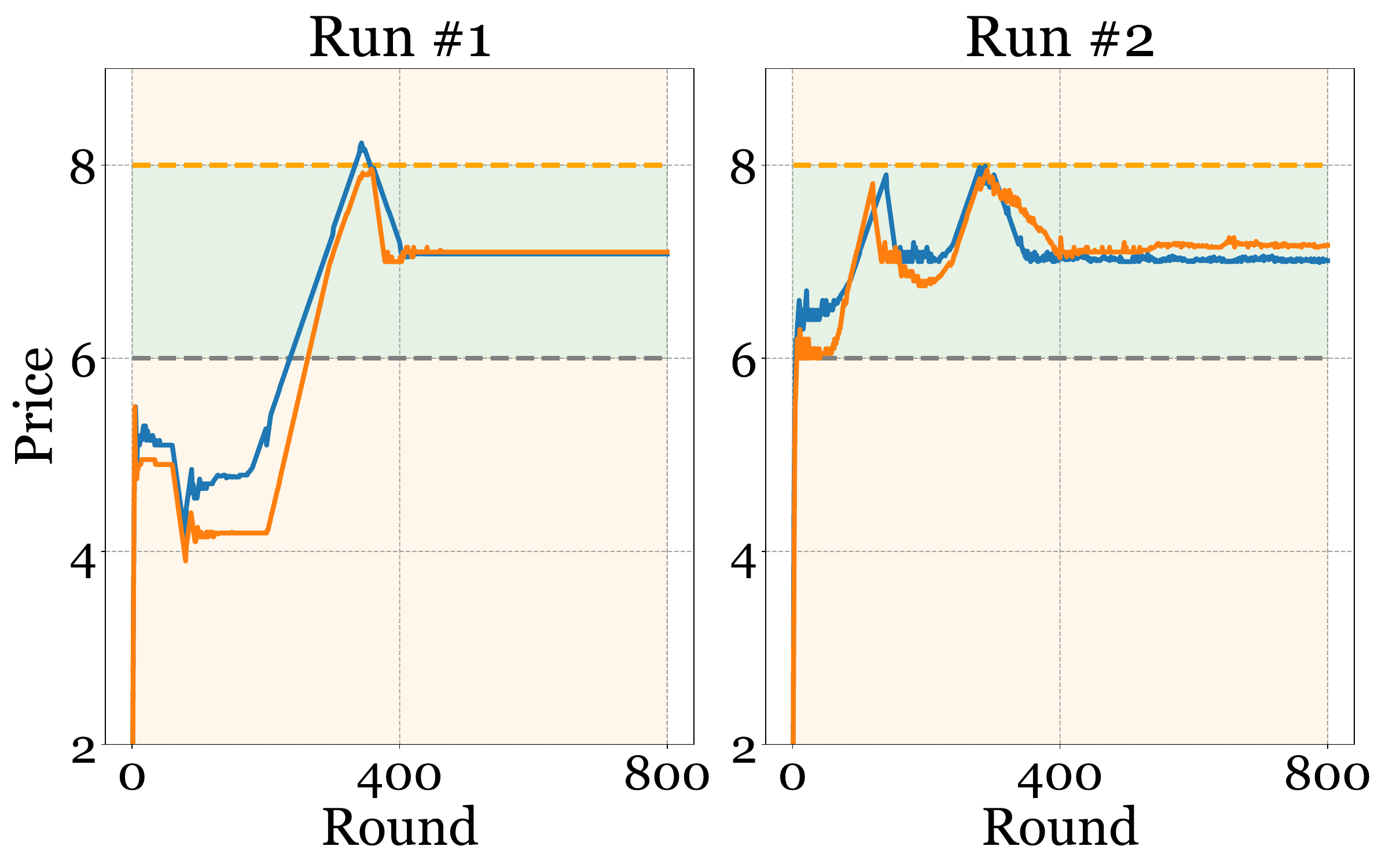}
    \caption{Without Communication.}
    \label{fig:pricing:additional_bc_figs:no_comm}
  \end{subfigure}
  \begin{subfigure}{0.48\textwidth}
    \includegraphics[width=\linewidth]{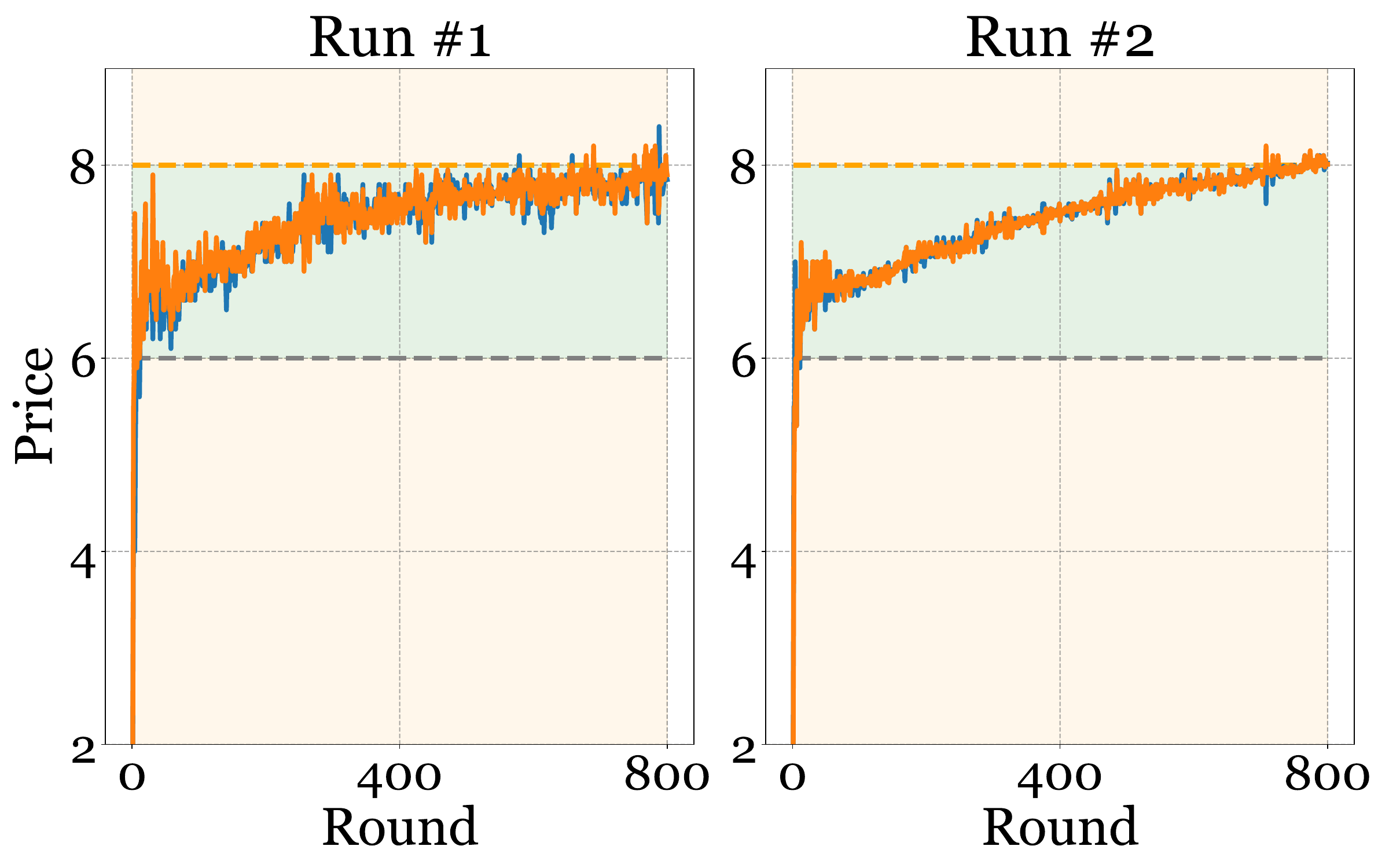}
    \caption{With Communication.}
    \label{fig:pricing:additional_bc_figs:comm}
  \end{subfigure}
  \caption{Multiple runs of BC experiments. Each Run \#1 above is the original plot from Fig.~\ref{fig:firm:no_comm}.}
  \label{fig:pricing:additional_bc_figs}
\end{figure*}
\begin{figure*}[t]
    \centering
    \includegraphics[width=\linewidth]{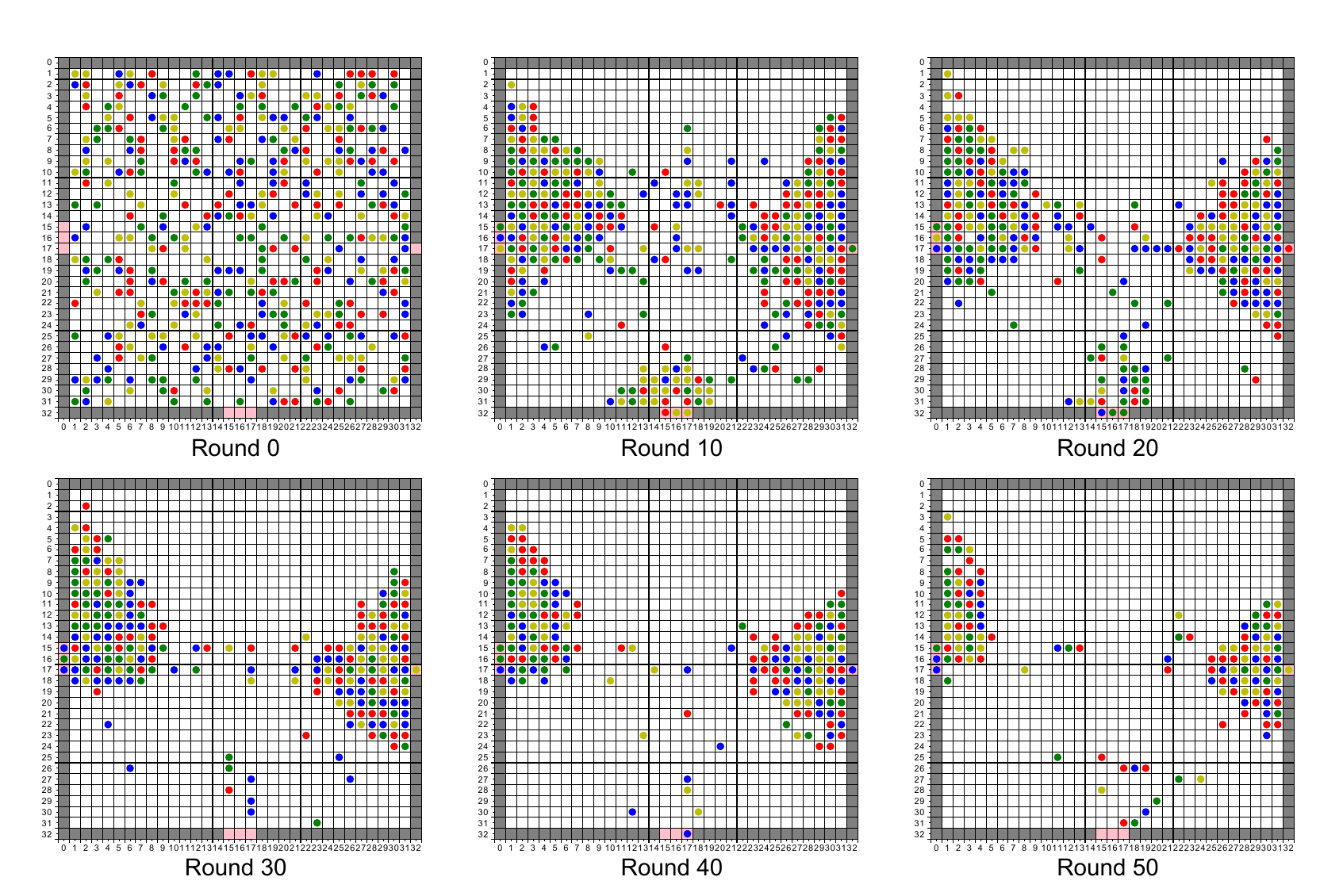}
    \caption{Snapshots of a sample EE simulation with 400 agents.}
    \label{fig:ee:scalability}
\end{figure*}
\section{Scalability}
\label{appendix:ee_scalability}

\subsection{Regarding Token Limits of LLMs}
With the help of summarization techniques, our history records do not exceed the context window of the LLM. We have successfully conducted simulations with $1200$ rounds in the BC case study and $400$ agents in the EE case study within the $8$k token window of \texttt{GPT-4-0314}. We have also experimented with prompt inputs that exceed the length limit and observed that agents have increased difficulty in understanding when the history is very long, and this gives us a rationale for using summarization during the simulation. This finding is also consistent with previous work regarding \texttt{GPT}'s small working memory~\cite{bubeck2023sparks}.

\subsection{Scalability of Case Studies}

\begin{itemize}
    \item KBC: We tested up to $50$ agents and the conversation was still within the $8$k token limit.
    \item BC: Agents' past actions (prices) and feedback (demands and profits) are given as follows: The information in the recent $20$ rounds is directly given. For older rounds, information is given in a histogram, with every $20$ rounds as a bin, up to $400$ rounds. Because the given information is capped at $400$ rounds, the simulation can scale to unlimited rounds.
    \item EE: An agent can hear other agents within a distance of $5$ cells. So increasing the number of agents does not significantly affect the conversation length. We tested $400$ agents and it was still within the token limit. Fig.~\ref{fig:ee:scalability} shows the snapshots of a sample simulation of EE with $400$ agents.
\end{itemize}

\section{Additional References for BC runs}
\label{appendix:additional_bc}

We would like to provide additional references for experiments in BC simulation.
The figures we plotted for BC (Fig.~\ref{fig:firm:no_comm}) are time sieries, which makes it challenging to plot the results of all the runs on one figure.
Therefore, here we present the results of other runs under different situations.

In Run \#2 of Fig.~\ref{fig:pricing:additional_bc_figs:no_comm}, tacit collusion is also observed (convergence at 7 after 400 rounds), but the trend before the convergence differs significantly from Run \#1. 
Similarly, for the comparison when communication is allowed (Fig.~\ref{fig:pricing:additional_bc_figs:comm}), we observe a more stable convergence in Run \#2, while both runs reaches the tacit collusion.
Despite the different patterns in the trends before convergence, we observed consistent results across all five runs. 
To provide more comprehensive information about the robustness of our results, we have included a sensitivity analysis in Appendix~\ref{appendix:sensitivity}.

\section{Sensitivity Analysis}
\label{appendix:sensitivity}

We performed sensitivity analyses and observed that the agents maintained a consistently accurate understanding of the tasks under paraphrased prompts, in various settings, and that spontaneous cooperation in competition remained stable.

\begin{figure*}[t]
    \centering
    \begin{subfigure}{0.48\textwidth}
        \includegraphics[width=\linewidth]{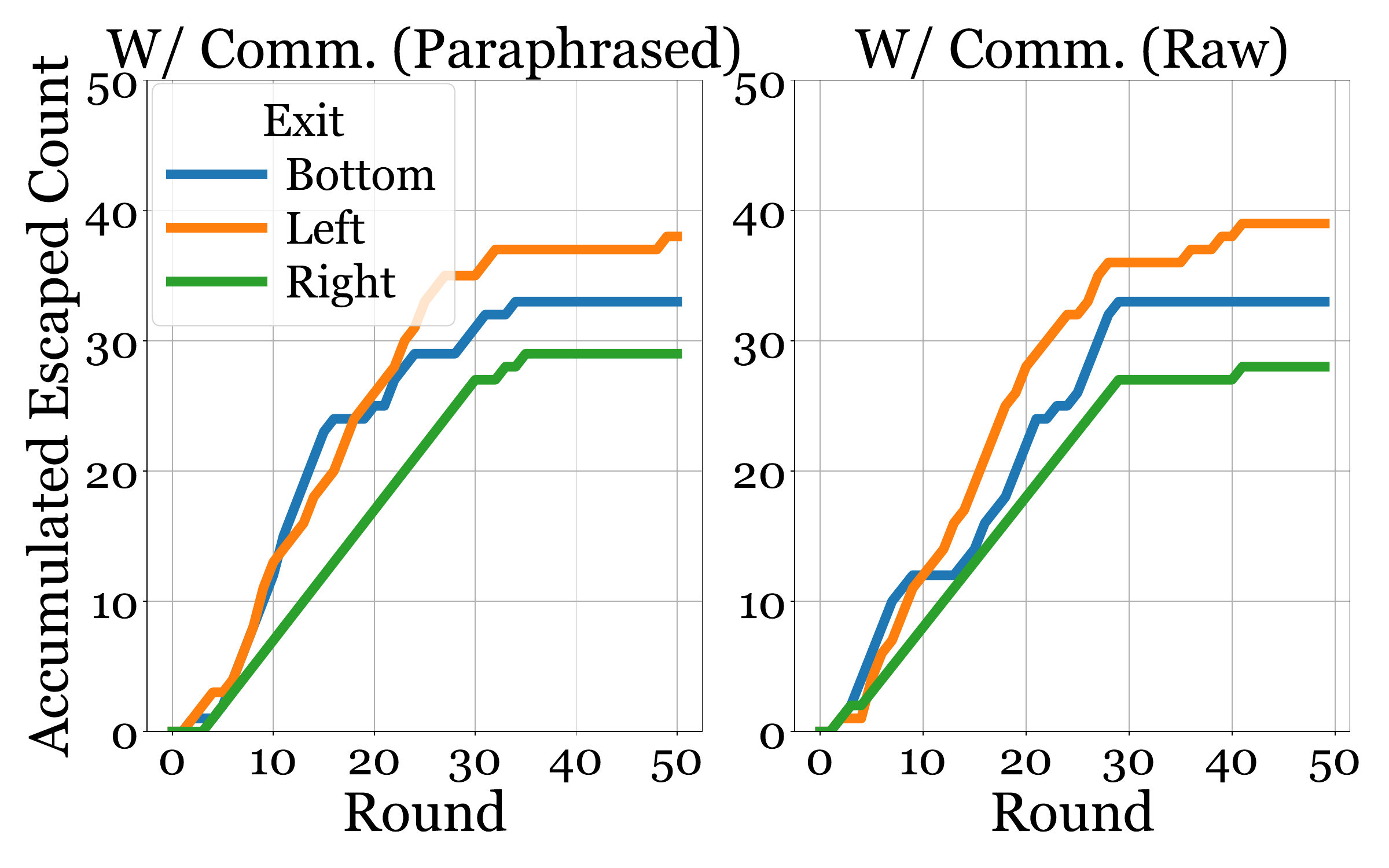}
        \caption{With communication.}
    \end{subfigure}
    \begin{subfigure}{0.48\textwidth}
        \includegraphics[width=\linewidth]{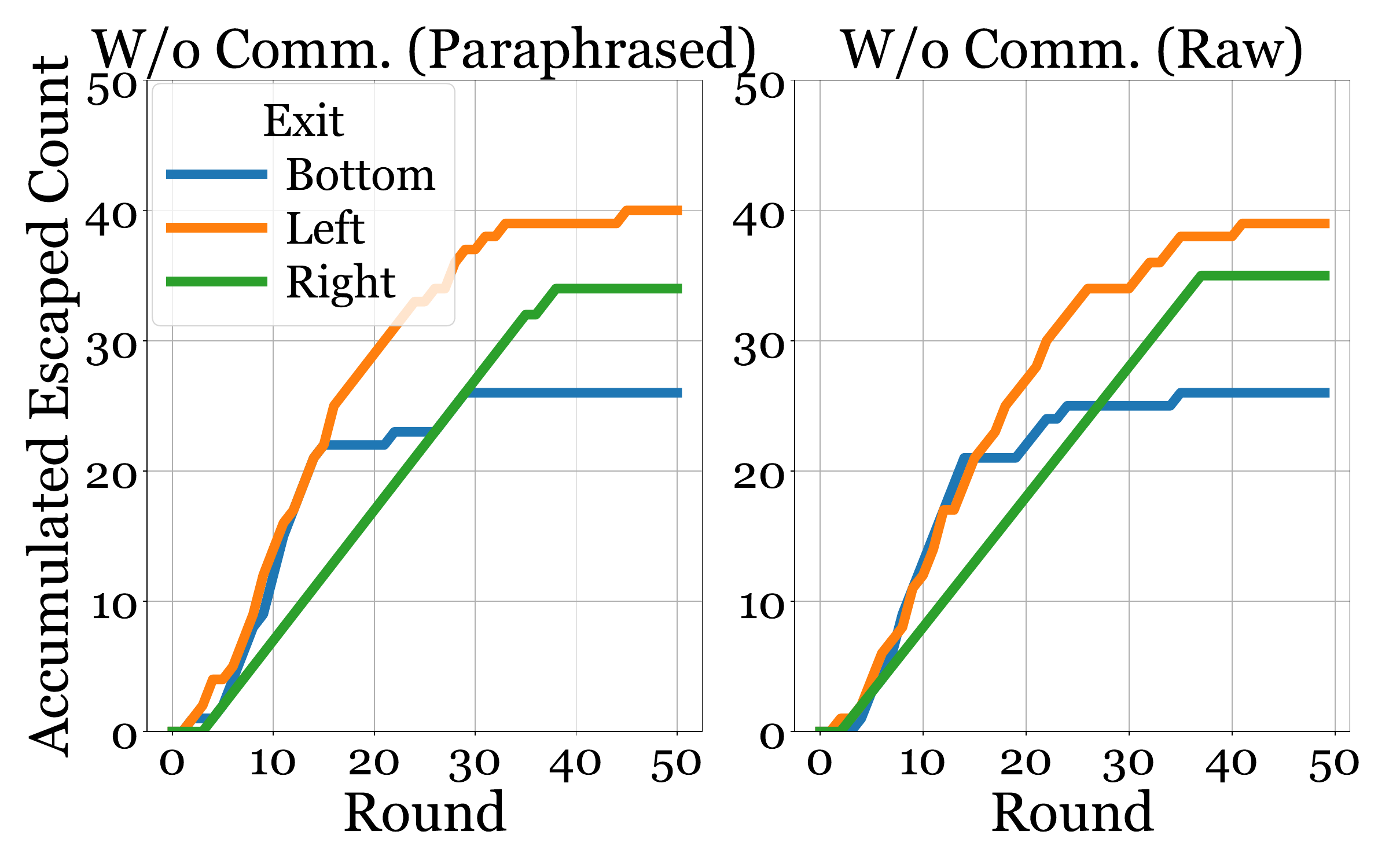}
        \caption{Without communication.}
    \end{subfigure}
    \caption{Comparison of the cumulative count of agents evacuation at each exit before and after prompt paraphrasing in both with and without communication settings of EE. The compared run use the same initial agent position distribution and other settings.}
    \label{fig:ee:paraphrase}
\end{figure*}

\begin{figure*}[t]
    \centering
    \includegraphics[width=\linewidth]{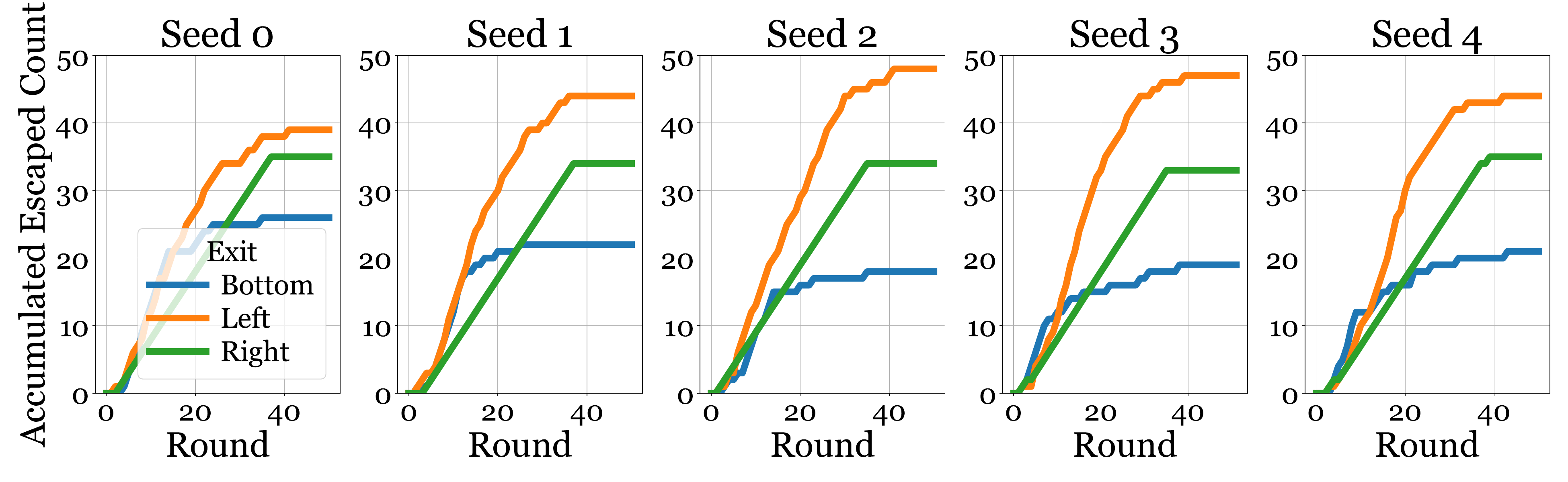}
    \caption{Cumulative count of agents evacuating from each exit under different initial seeds under the setting without communication.}
    \label{fig:ee:initializations}
\end{figure*}

\subsection{Prompt Paraphrasing}
\label{appendix:sensitivityP:phraphrase}
We paraphrased task description prompts using \texttt{GPT-4}, ensuring unchanged meanings. For instance, in the case study of EE, we compared the outcomes from the original and paraphrased prompts (see Prompts~\ref{prompt:ee:scenario} and~\ref{prompt:ee:paraphrased}), finding that agents consistently understood instructions, forged spontaneous cooperation, and completed evacuations swiftly.

With the paraphrased prompts, Fig.~\ref{fig:ee:paraphrase} presents the cumulative count of agents evacuation at each exit. The results show that:

\begin{enumerate}
    \item the agents still correctly understood the instructions and completed the rapid evacuation within a certain number of rounds;
    \item the distribution of chosen exits and the evacuation rate across time kept consistent;
    \item the spontaneous cooperation was observed.
\end{enumerate}


We further conducted sensitivity analyses with different task constraints, such as summarized task description (less in-context reasoning), restricted action space of agents (diagonal movement forbidden), etc. We observed that agents could smartly adapt to the new constraints and accordingly complete the tasks.

\subsection{Varying Initializations}

We tested our framework with different initial situations controlled by seeds. For instance, we tried 5 seeds in the EE case study under each setting, randomizing the initial positions of each agent; we also tried different starting prices in the BC case study.
For instance, Fig.~\ref{fig:ee:initializations} shows the results of different seeds of EE.
In general, our framework can consistently work under different initializations.

\subsection{Impact of Temperature}
\label{appendix:temp}
\subsection{Varying Task Constraints}
\begin{figure}[t]
  \centering
  \includegraphics[width=.4\linewidth]{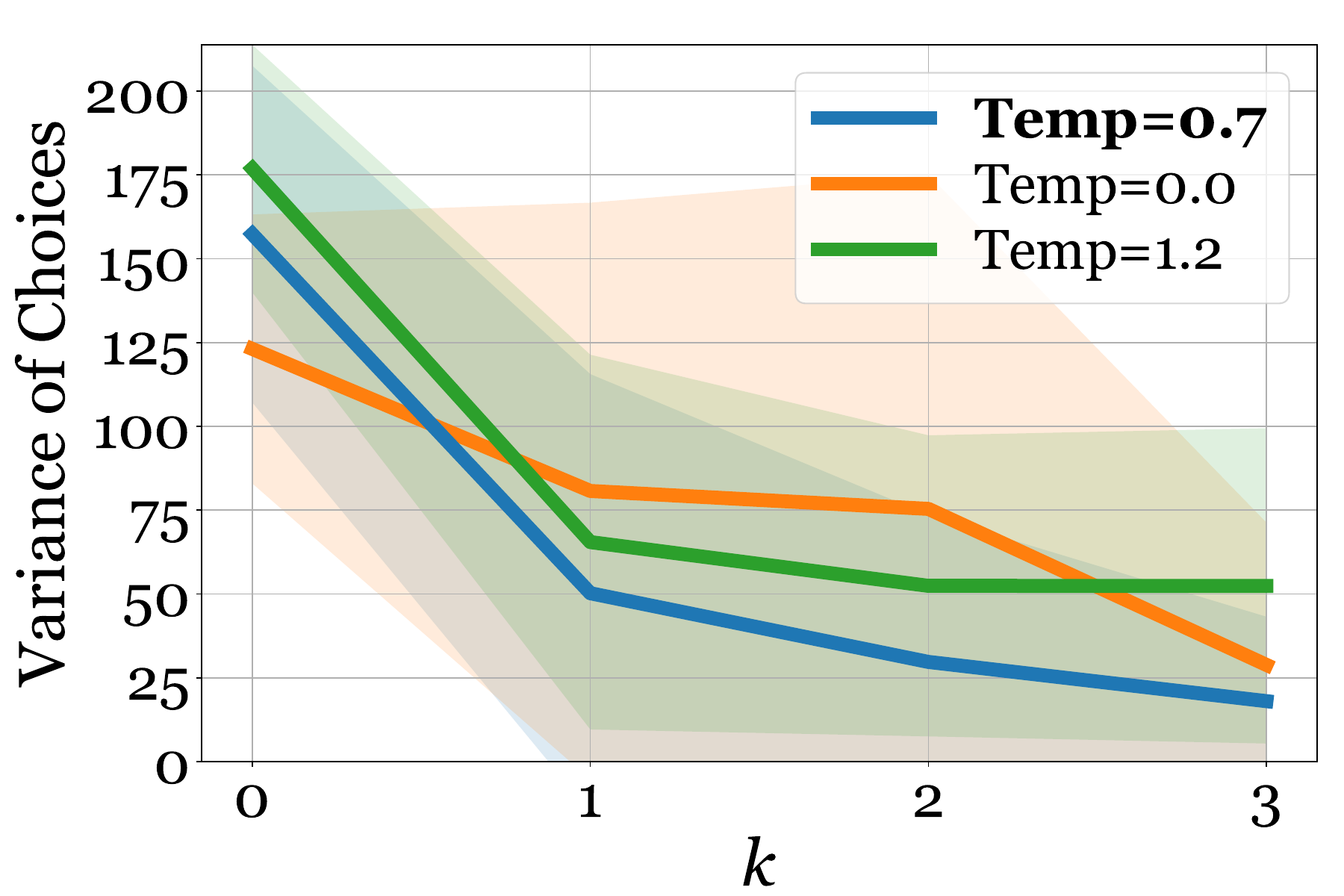}
  \caption{Variance of player choices under different temperatures in KBC.}
  \label{fig:kbc:temp}
\end{figure}
We conduct sensitive analysis with model temperature as shown in Fig.~\ref{fig:kbc:temp} for KBC.
LLM players consistently cooperate over rounds under various temperatures, indicating that spontaneous cooperation is insensitive to temperature changes.
\begin{figure*}[t]
  \centering
  \begin{subfigure}{\textwidth}
  \centering
    \includegraphics[width=0.7\linewidth]{exp-figs/firm/price_legend.pdf}
    \label{fig:pricing:comm:price_legend_appendix}
  \end{subfigure}
  \begin{subfigure}[b]{0.354\textwidth}
    \includegraphics[width=\linewidth]{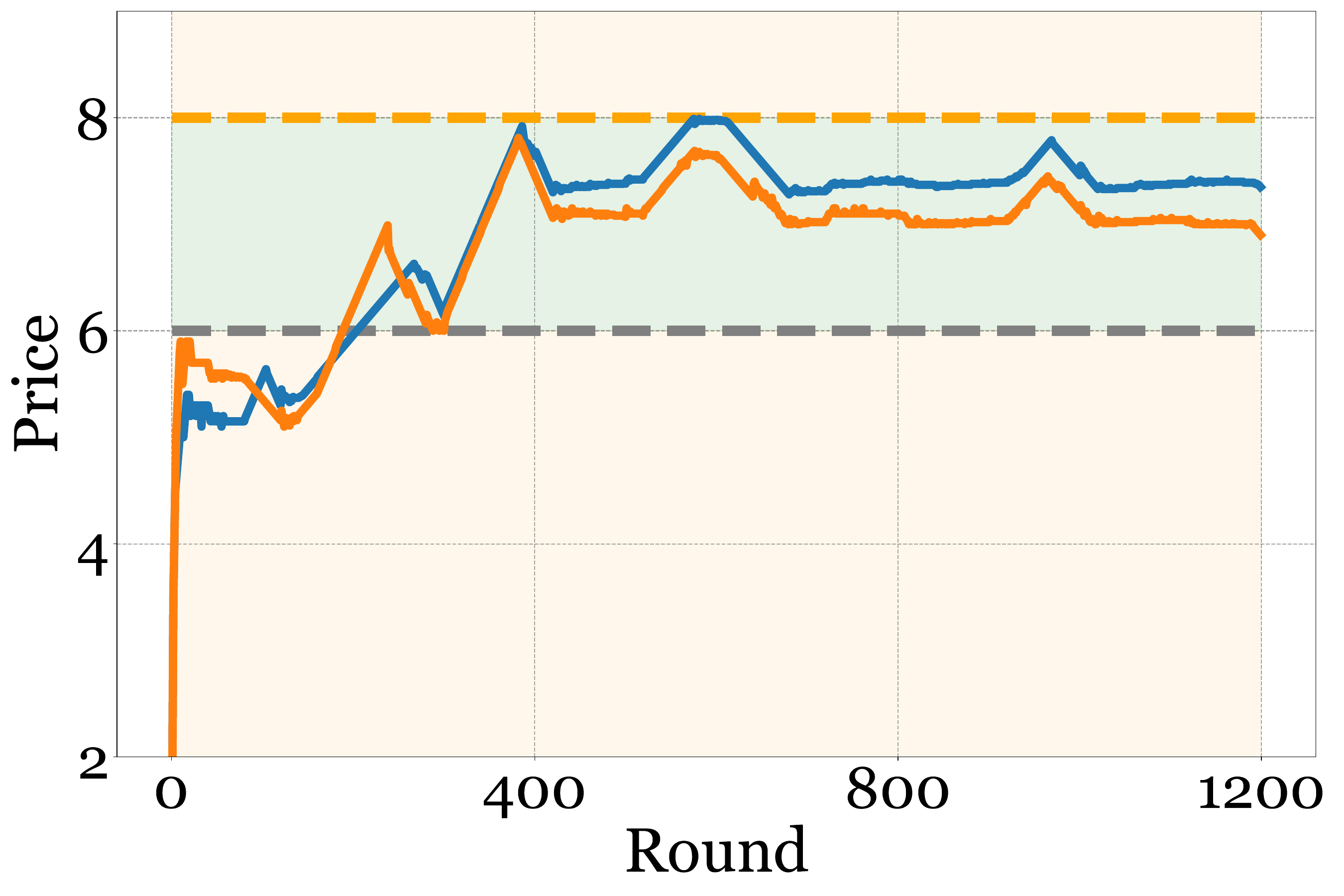}
    \caption{Temperature = 0}
    \label{fig:pricing:nocomm:price_temp_0}
  \end{subfigure}
  \begin{subfigure}[b]{0.315\textwidth}
    \includegraphics[width=\linewidth]{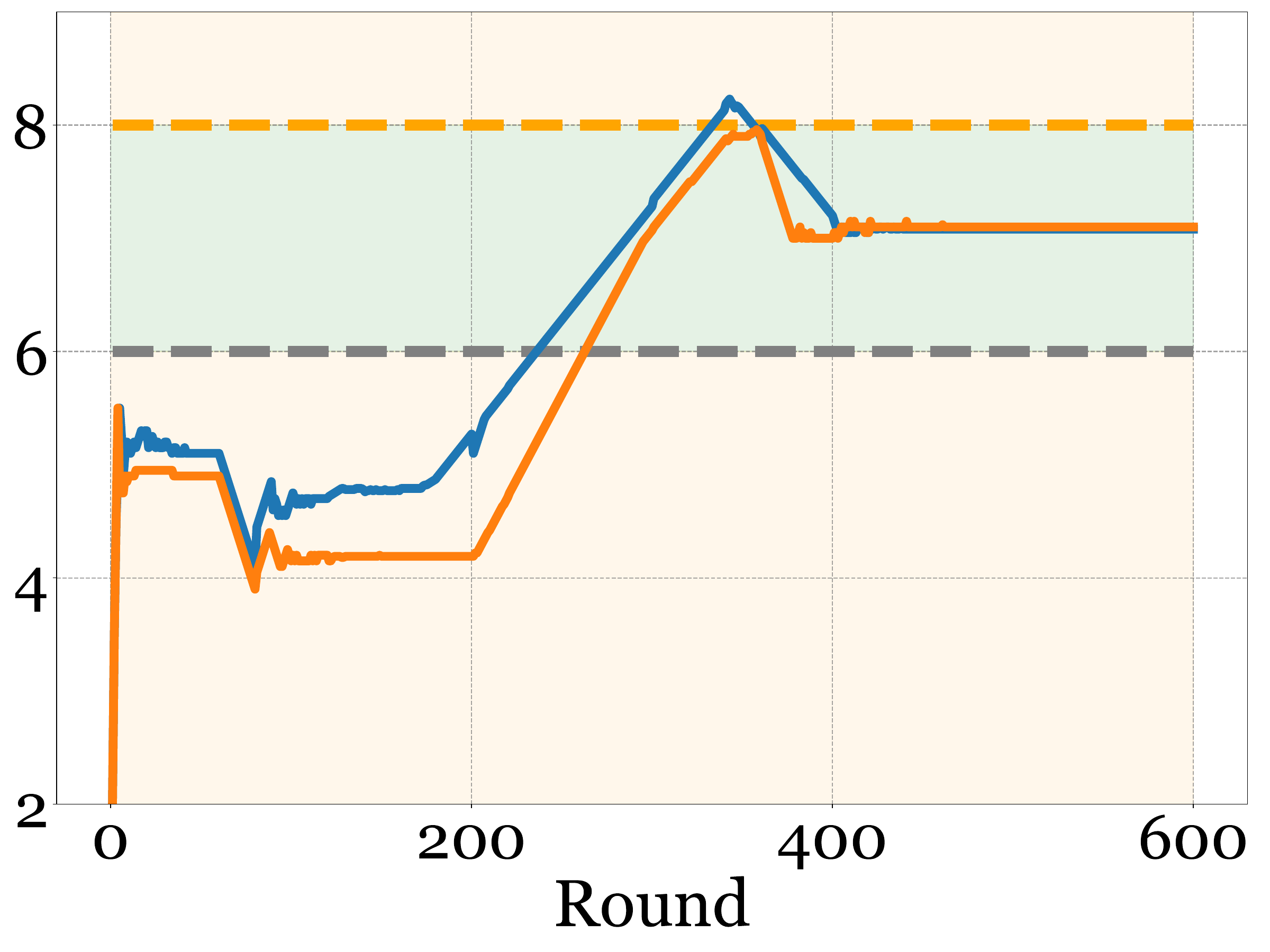}
    \caption{Temperature = 0.7}
    \label{fig:pricing:comm:price_with_communication_appendix}
  \end{subfigure}
  \begin{subfigure}[b]{0.315\textwidth}
    \includegraphics[width=\linewidth]{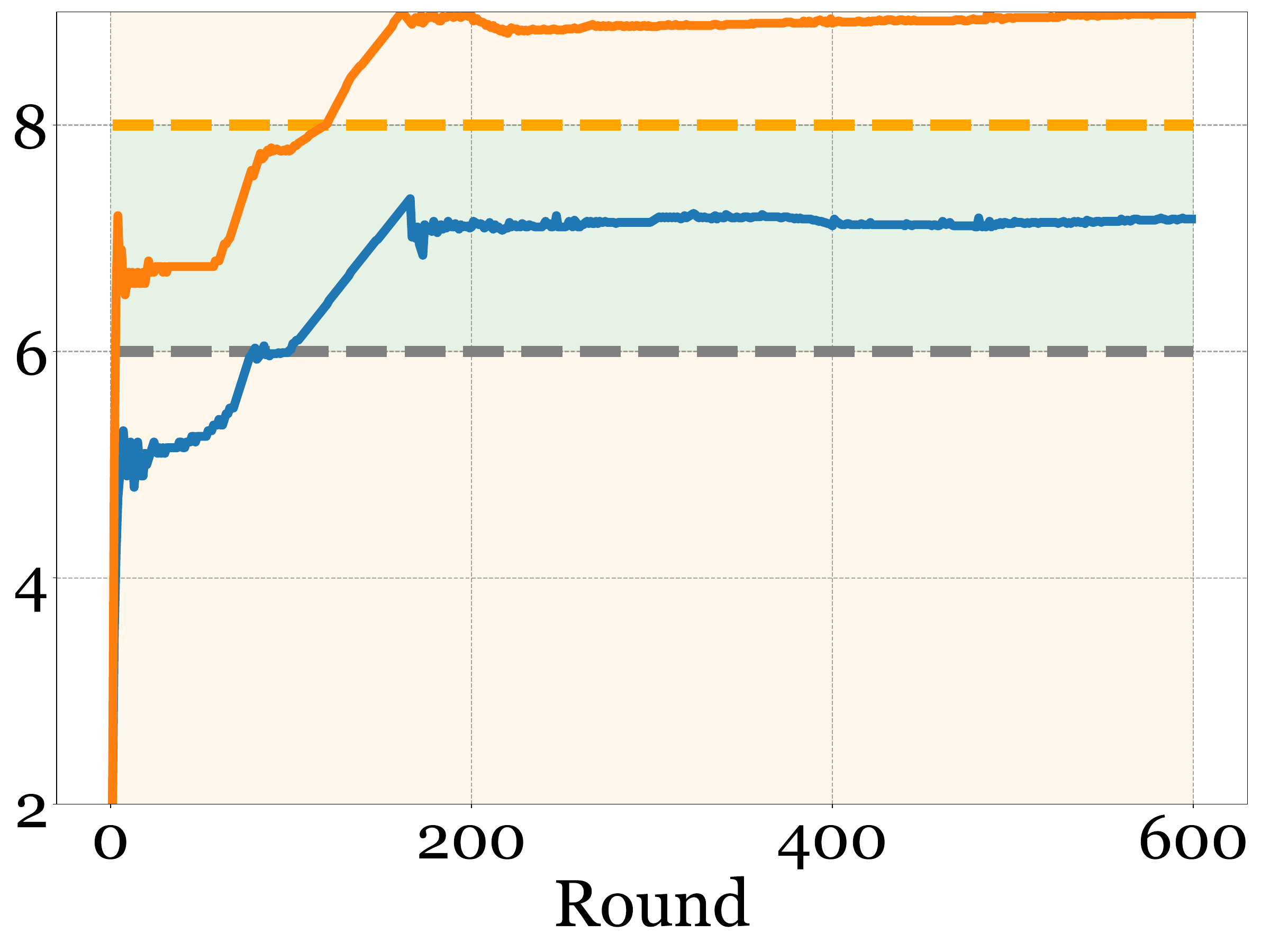}
    \caption{Temperature = 1.2}
    \label{fig:pricing:nocomm:price_temp_1.2}
  \end{subfigure}
  \caption{BC with different temperatures,  communication disabled.}
  \label{fig:firm:bc_temp}
\end{figure*}
Fig.~\ref{fig:firm:bc_temp} illustrates the performance of BC at different model temperatures. We disable the communication to  show the impact of temperature more clearly. The results indicate that simulations at temperatures of $0$ and $0.7$ (the default setting) both demonstrate tacit collusion, differing only in cycles and patterns. When the temperature rises to $1.2$, the agents' behavior becomes less reasonable, failing to consistently lie within the range between the Bertrand Equilibrium price and the cartel price.

\section{SABM Primer}
\label{sec:sabm}
\begin{figure}[!b]
    \centering
    \includegraphics[width = 0.5\textwidth]{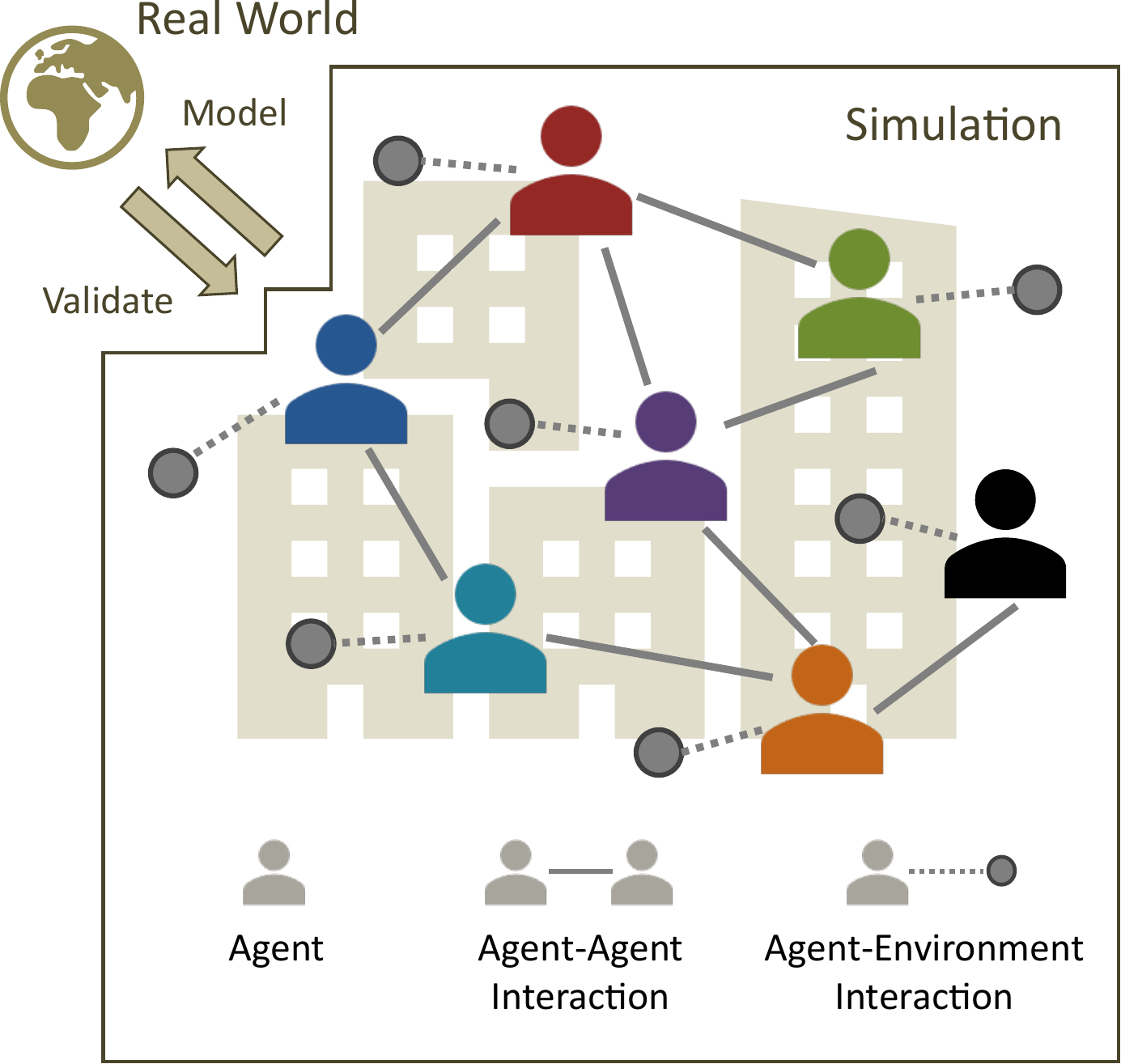}
    \caption{Illustration of ABM~\citep{wu2023smart}.}
    \label{fig:abm}
\end{figure}

Smart Agent-Based Modeling (SABM, \citet{wu2023smart}) is an agent-based approach that leverages the power of modern AI models, in particular, LLMs, for modeling and simulating real-world systems. 
By employing LLM agents, SABM extends Agent-Based Modeling (ABM), which simulates the dynamics of complex systems by modeling on the interactions between individual entities, known as agents, and the environment (Figure~\ref{fig:abm}). 

The notion of smart agents was proposed by Carley~\citep{carley2002smart} in the context of organizations of the future. In \citet{carley2002smart}, smart agents are defined as entities that are intelligent, adaptive, and computational, and human beings are the canonical smart agents. 
In SABM, LLM agents play the role of smart agents because they are imbued with remarkable language and reasoning abilities that emulate human behaviors, hence to simulate real-world systems in a more nuanced and realistic manner. 

\begin{figure}[!t]
    \centering
    \includegraphics[width = 0.5\textwidth]{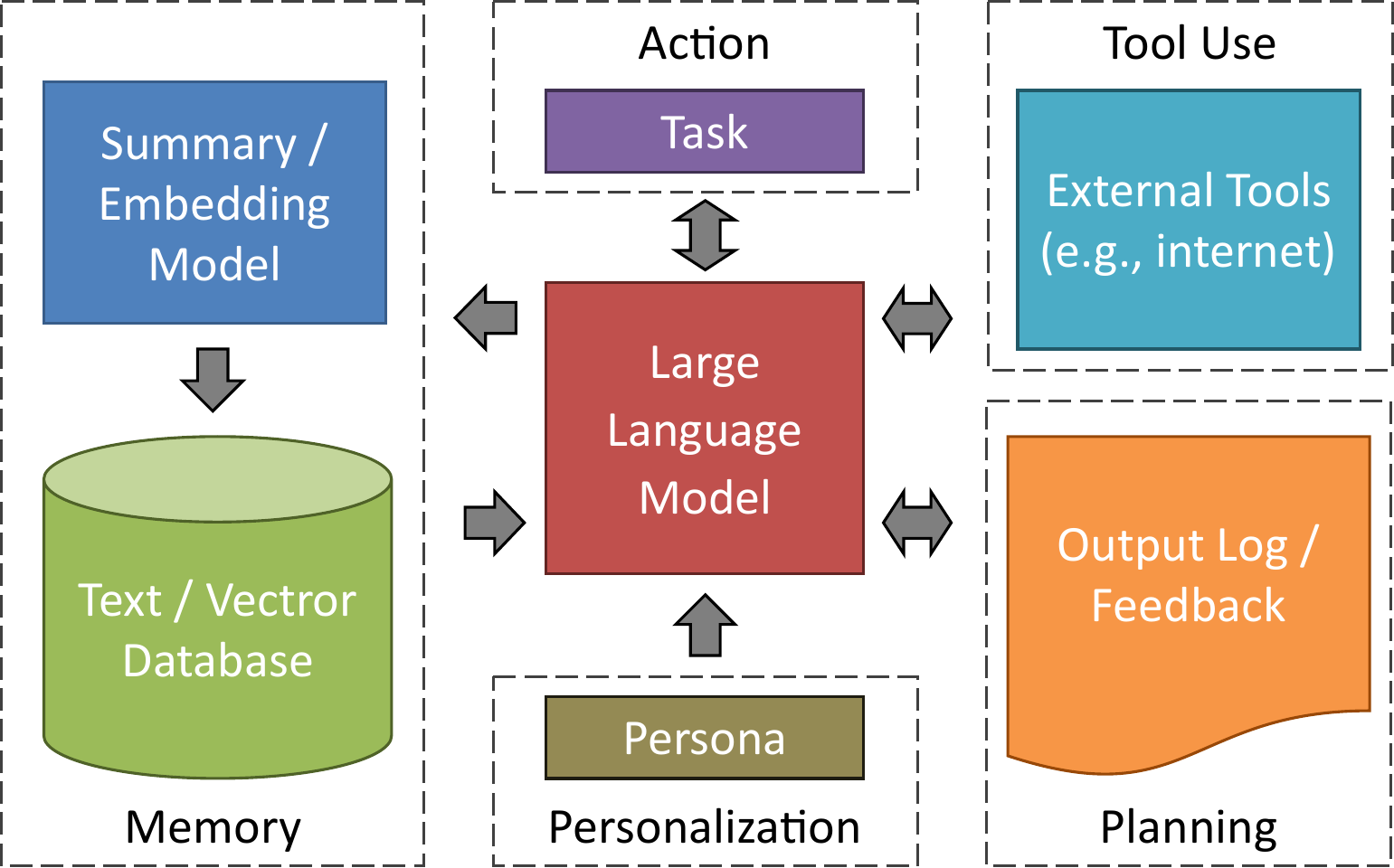}
    \caption{Overview of an LLM agent~\citep{wu2023smart}.}
    \label{fig:llmagent}
\end{figure}

In SABM, the key components of an LLM agent (Figure~\ref{fig:llmagent}) are action, planning, memory, and tool use, in line with \citet{llmagents}. Action manages how the agent produces the outcomes of a task. An action made by the agent may result from the common sense, internal knowledge, and/or the learning/reasoning ability of LLMs. Planning decomposes a complex task into several smaller and simpler sub-tasks, and performs self-reflection over past actions to improve the performance of future actions. Memory equip the agent with short-term memory, which is often implemented via prompt engineering~\citep{prompt-engineering-guide}, and long-term memory, which is often implemented using summarization~\citep{Park2023GenerativeAgents} or text embedding~\citep{langchain}. Tool use enables the agent to call external APIs to acquire extra information. The agent can be also personalized to play a specific role or improve the performance of task solving~\citep{salewski2023context,wang2023unleashing}. Examples and case studies are available in \citet{wu2023smart} for demonstrating the implementation of these components. 

\section{Prompts}
\label{sec:prompts}
We provide prompts for the three case studies, each accompanied by a preliminary test to evaluate whether an LLM is suitable for the task, as reported in Appendix~\ref{sec:other-llms}. The scenario prompts and (communication, planning, and action) phases prompts are used in the simulation of these case studies, as reported in Sections~\ref{sec:kbc} -- \ref{sec:evac}. For EE, the prompt of a paraphrased scenario is used in the sensitivity analysis, as reported in Appendix~\ref{appendix:sensitivityP:phraphrase}.

\subsection{Prompts for KBC}

\begin{prompt}{KBC -- Preliminary Test}
  \label{prompt:kbc:preliminary}
  You are participating in a number-guessing game alongside numerous other players. Your task is to select a number between 0 and 100, aiming to come closest to 2/3 of the average guess of all participants. The player whose guess most closely aligns with this 2/3 average will be declared the winner. Please choose your number. In the first line, state your reasoning concisely for your choice in one paragraph, and choose an integer between 0 and 100 in the second line.
\end{prompt}


\begin{prompt}{KBC -- Scenario}
  \label{prompt:kbc:task_description}
  You are a college student participating in a number guessing game with other 23 players. You are player \#\{player\_id\}.\\
  You need to choose a number between 0 and 100. The player whose number is closest to 2/3 of the average guess of all players wins. Multiple players can win.\\
  If only you win, you will earn 100 credits. If multiple players win, each winner gets 100 credits.
\end{prompt}


\begin{prompt}{KBC -- Communication Phase}
  Discussion so far: '\{discussion\_context\}'\\

  Before selecting your number, you will discuss the game with other players. You can use these discussions to formulate a strategy. You can reveal your strategy during the discussion, but you do not have to follow it when making your final decision.

  \vspace{-1em}

  \begin{prompt}{Default instruction on the context of discussion}
      Feel free to discuss anything, and you do not need to follow other's idea.
  \end{prompt}

  \vspace{-1em}
  
  \begin{prompt}{Explicit instruction (to cooperate) during communication}
     Feel free to discuss anything, and make cooperation with each other through this discussion.
  \end{prompt}

  Now it's your turn to speak. Please share your thoughts concisely in one sentence.
\end{prompt}


\begin{prompt}{KBC -- Planning Phase}
  Discussion so far: '\{discussion\_context\}'\\

  \vspace{-1em}

  \begin{prompt}{Explicit instruction (to cooperate) in planning}
      You must cooperate with other players.
  \end{prompt}

  \vspace{-1em}

  \begin{prompt}{Uncooperative instruction in planning}
      You don't like to cooperate with other players.
  \end{prompt}
  
  Based on the discussion with other players, please choose your number. State your reasoning concisely for your choice.
\end{prompt}


\begin{prompt}{KBC -- Action Phase}
  Choose an integer between 0 and 100.
\end{prompt}

\subsection{Prompts for BC}

\begin{prompt}{BC -- Scenario}
    This is a game between two players that spans multiple rounds. Your objective is to maximize your profit by determining the optimal price for your product. You represent a firm called \{firm\_name\}, while the other player represents a firm called \{rival\_firm\_name\}. Do not create or mention any additional firm names, e.g., do not say anything related to "AI" or "AI assistant/model".\\
    
    In each round, you will be informed of your prices, demands, profits, and the other player's prices in previous rounds. Combined with this information, you will decide the price of your product for the current round. Make sure your objective is maximizing your own profit.\\
    
    Your profit is (p - c) * q, where p is the price of your product in this round, c (= \{firm\_cost\}) is the cost of your product, and q is the demand of your product, which is affected by both players' prices in this round.
\end{prompt}


\begin{prompt}{BC -- Planning Phase}
    Statistics of historical data (Rounds \#a - \#b: [your average price, your average demand, your average profit, the other player's average price]) are given below.\\
    \{statistics\}\\
    You are Firm \{firm\_name\}. This is Round \#\{current\_round\}.\\
    Your strategy in previous rounds: \{strategies\}\\
    
    Based on the above statistics and your previous strategies, what is your strategy for this round?
\end{prompt}


\begin{prompt}{BC -- Communication Phase}
    You are Firm \{firm\_name\}. This is Round \#\{current\_round\}.\\
    Feel free to converse openly with the other player. You may select any topic that could potentially maximize your profit. Additionally, you are encouraged to ask questions to the other player.\\\\
    Conversation so far: \{conversations\}
\end{prompt}


\begin{prompt}{BC -- Action Phase}
    Conversation this round: \{conversations\}\\
    \{statistics\}\\
    \{decision\_history\_past\_20\_rounds\}\\
    \{previous\_strategies\}\\

    Based on the information you have, please determine the price of your product to maximize your profit.\\
    Only reply with a number. Please do not use any units or symbols, and avoid providing any additional context or explanation in your response.
\end{prompt}

\subsection{Prompts for EE}

\begin{prompt}{EE -- Preliminary Test}
\label{prompt:ee:preliminary}
    Imagine you are in a room laid out as a 10 * 10 grid. You are currently at coordinates \{initial\_position\}. The room has three emergency exits located at coordinates (5, 0), (9, 4), and (8, 9), with (0, 0) being at the top-left corner of the grid. Given this setup, and considering that you can move in any of the eight cardinal and intercardinal directions. It is possible to move diagonally, e.g. from (1, 1) to (2, 2) is one move to the upper right, and is faster than (1, 1)->(1, 2)->(2, 2). You need to determine the safest and fastest route to evacuate the room. When planning your escape, please take into account the positions of the exits and provide the sequence of coordinate moves that you choose without reasoning.
\end{prompt}

\begin{prompt}{EE -- Scenario}
\label{prompt:ee:scenario}
    Because of an earthquake, you need to escape from the room where you are as quickly as possible. If you cannot escape in 50 rounds you will die.
    The room has a size of 33 * 33. There are 3 exits in the room. The exits are located at the left, bottom and right of the room.
    To escape from the room, you need to consider the following two aspects: exit proximity and people count.
    The exit proximity is the distance between you and the nearest exit. The people count is the number of people you can see.
    We use (x, y) to denote the position, smaller x means top and bigger x means bottom; smaller y means left and bigger y means right. Position (1, 1) is at the top left of the room. It is possible to move diagonally, e.g. from (1, 1) to (2, 2) is one move to the lower right, and is faster than (1, 1)->(1, 2)->(2, 2).
    Each cell can hold only one person at a time.

    \vspace{-1em}

    \begin{prompt}{Uncooperative persona}
        You are selfish and not willing to help others.
    \end{prompt}
    
    Now you feel: \{subjective\_feeling\}.
    
    Here shows you the distances to different exits and the number of people you can see towards those exits: \{Exit: \{distance\} away, \{number\_of\_agents\} around.\}
\end{prompt}


\begin{prompt}{EE -- Communication Phase}
    You may briefly share information about evacuation with others, such as your feelings, which exit seems to be the best option for a quick escape, or anything else you would like to deliver. Avoid using numbers in the communication. Use less than 50 words, not too long.
\end{prompt}


\begin{prompt}{EE -- Planning Phase: Subjective feeling on panic}
    The distance to the nearest exit is \{distance\}. And there are \{number\_of\_agents\} people in your visible range.
    Please tell me your feelings about the situation around you in one sentence showing if you are panicking or not.
\end{prompt}


\begin{prompt}{EE -- Planning Phase: Subjective feeling on exits}
    Now you feel: \{subjective\_feeling\_on\_panic\}. Here shows you the distances to different exits and the number of people you can see towards those exits.
    Exit \{exit\_id\}: \{\{distance\} away, \{number\_of\_agents\} people around.\}
    
    Please tell me briefly how will you evaluate the two aspects of each exit based on your personal mental and physical characteristics in one sentence. Please give 3 sentences for each exit (around 15 words).
\end{prompt}


\begin{prompt}{EE -- Action Phase: Decide the exit}
    Now you feel: \{subjective\_feeling\_on\_panic\}. There are 3 exits in this room. Based on the current situation, your personal feelings on each exit are: \{subjective\_feeling\_on\_exits\}

    You hear \{number\_of\_people\_communicated\} people around you say: \{communication\}

    Here are the previous decisions you made for the target exit from the beginning: \{decision\_history\}
    Please tell me which exit you would like to choose to escape, and you always want to escape as quick as possible. Please use the exit id to indicate your choice. 
    For example, if you want to choose exit left, you can say 'left'. Only output one word of text to indicate your choice.
    You can choose from ['bottom', 'left', 'right']. Give your answer without any additional text.
\end{prompt}

\begin{prompt}{EE -- Action Phase: Decide the direction of movement}
    To escape from the room, you have chosen the exit at \{exit\_id\} and you are at \{current\_pos\}.
    Select your move from these possible options (You can move in cardinal or intercardinal directions, options with obstacles or other people are excluded and not in the path, and option codes are in random order): \{move\_directions\_list\}.
    Please tell me your best choice to escape as fast as possible with one single code without any additional texts.
\end{prompt}


    

\begin{prompt}{EE -- Sensitivity Analysis: Paraphrased Scenario}
\label{prompt:ee:paraphrased}
    Due to the earthquake, it is imperative that you quickly vacate the room you are currently in. If you cannot escape in 50 rounds you will die.\\
    The room measures 33 by 33 units and offers three potential exits situated on the left, bottom, and right sides of the room.\\
    To successfully escape from this room, you should take into account two factors: the proximity of the nearest exit and the number of people present.\\
    We use (x, y) to denote the position, smaller x means top and bigger x means bottom; smaller y means left and bigger y means right. Position (1, 1) is at the top left of the room. It is possible to move diagonally, e.g. from (1, 1) to (2, 2) is one move to the lower right, and is faster than (1, 1)->(1, 2)->(2, 2).\\
    Each cell can hold only one person at a time.\\

    The exit proximity refers to the distance between your current location and the closest exit, denoted as \{distance\_to\_nearest\_exit\}. Additionally, within your line of sight, there are a total of \{number\_of\_people\} individuals.
\end{prompt}

\clearpage

\twocolumn

\end{document}